\documentclass[10pt,twocolumn,letterpaper]{article}

\usepackage[table,dvipsnames,svgnames]{xcolor}
\usepackage{cvpr}              % To produce the CAMERA-READY version
\usepackage{graphicx}
\usepackage{nicematrix}
\usepackage{pifont}% 
\usepackage{multirow}
\usepackage{cuted}
\usepackage{capt-of}
\usepackage{comment}
\usepackage[toc,page,header]{appendix}
\usepackage{minitoc}
\newcommand{\cmark}{\ding{51}}%
\newcommand{\xmark}{\ding{55}}%

\definecolor{cvprblue}{rgb}{0.21,0.49,0.74}
\usepackage[pagebackref,breaklinks,colorlinks,allcolors=cvprblue]{hyperref}

\newcommand{\az}[1]{\textcolor{blue}{[az: #1]}}
\newcommand{\methodName}{\textsc{CountGD++}}

\def\paperID{6497} % *** Enter the Paper ID here
\def\confName{CVPR}
\def\confYear{2026}

\title{CountGD++: Generalized Prompting for Open-World Counting}

\author{Niki Amini-Naieni\\
Visual Geometry Group (VGG)\\
University of Oxford, UK\\
{\tt\small nikian@robots.ox.ac.uk}
\and
Andrew Zisserman\\
Visual Geometry Group (VGG)\\
University of Oxford, UK\\
{\tt\small az@robots.ox.ac.uk}
}

\begin{document}
\maketitle
\doparttoc % Tell to minitoc to generate a toc for the parts
\faketableofcontents % Run a fake tableofcontents command for the partocs

\part{} % Start the document part
%\parttoc % Insert the document TOC
\begin{comment}
\begin{strip}\centering
\includegraphics[width=0.9 \textwidth]{figures/main_teaser.pdf}
 
\captionof{figure}{(a) The negative visual exemplar (white blood cell boxed in red) enables \methodName\ to differentiate between cells that have the same round shape as the object to count (red blood cell boxed in green) but are of a different color; (b) Pseudo-exemplars are automatically detected from text only and fed back to the model, improving the accuracy of the final count for objects, like unfamiliar fruits, that are challenging to identify given text only. 
\label{fig:teaser}}
\end{strip}
\end{comment}
\begin{abstract}
%Background: [Why are we having this conversation now] There are concerns that air pollution has increased the transmission and spread of COVID.  
The flexibility and accuracy of methods for automatically counting objects in images and videos are limited by the way the object can be specified. While existing methods allow users to describe the target object with text and visual examples, the visual examples must be manually annotated inside the image, and there is no way to specify what \textbf{not} to count. To address these gaps, we introduce novel capabilities that expand how the target object can be specified. Specifically, we extend the prompt to  enable what \textbf{not} to count to be described with text and/or visual examples, introduce the concept of `pseudo-exemplars' that automate the annotation of visual examples at inference, and extend counting models to accept visual examples from both natural and synthetic \textbf{external} images. We also use our new counting model, \methodName, as a vision expert agent for an LLM. Together, these contributions expand the prompt flexibility of multi-modal open-world counting and lead to significant improvements in accuracy, efficiency, and generalization across multiple datasets. Code is available at \href{https://github.com/niki-amini-naieni/CountGDPlusPlus/}{https://github.com/niki-amini-naieni/CountGDPlusPlus/}.

%Population: 

%Methods: We performed a systematic review of XX  databases. Studies were included based on X. This yielded X studies, of which Y were eligible and included in the final sample.

%Results: [key findings which you highlight in the conclusion come here]

%Conclusions: Our study found X. Future research is needed to Y. 

\end{abstract} 
\begin{figure}[h!]
  \centering
  \includegraphics[width=\linewidth]{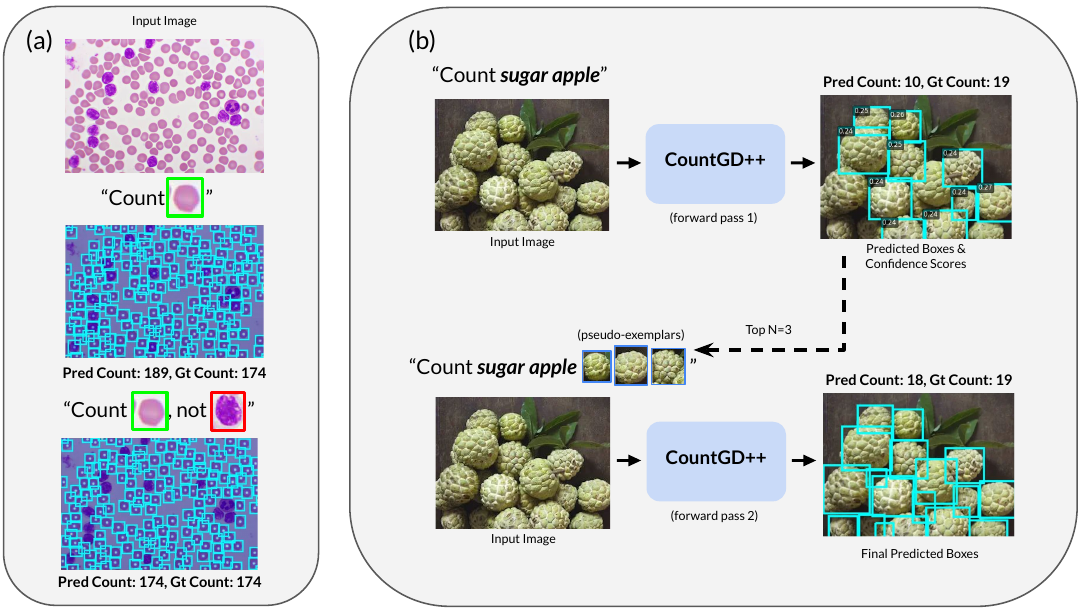}
  \vspace{-5mm}
  \caption{{\bf New capabilities of \methodName.} \textit{(a) Counting with Positive \& Negative Prompts:} The negative visual exemplar enables \methodName\ to differentiate between cells that have the same round shape as the object to count but are of a different appearance; \textit{(b) Pseudo-Exemplars: } Pseudo-exemplars are automatically detected from text only and fed back to the model, improving the accuracy of the final count for objects, like unfamiliar fruits, that are challenging to identify given text alone. 
\label{fig:teaser}}
\end{figure}
\section{Introduction}
\label{sec:intro}

% [Why are we having this conversation now]
% Object counting models have gotten very good as a result of expanding the prompts they can receive 

Recently, models for automatically counting any object in images have experienced widespread adoption~\cite{seed2025seed1_5vl, landingai, countse, demystifying_numerosity, countvid}. This is because advances in how the target object can be described, the \emph{prompt flexibility}, have led to significant improvements in counting accuracy and applicability across diverse problems~\cite{countgd}. However, this newfound attention has also revealed several limitations. 

The problems that state-of-the-art counting models can solve are restricted. The most accurate models struggle to distinguish between visually similar objects, require users to manually identify visual examples for every input image, and cannot accept visual examples from external sources. This means they cannot understand intuitive prompts such as the ones shown in \cref{fig:teaser}. These restrictions mean visually specifying objects can be extremely tedious, requiring the manual annotation of potentially thousands of images, and that prompts are unable to make subtle or nuanced distinctions. These limitations restrict counting models from being applied to many real-world problems such as counting different blood cells for medical diagnosis in medicine, measuring the formation rate of growing crystals in x-ray videos to develop more sustainable materials, and distinguishing between ripe and unripe fruits in agriculture.

%\az{Need to revise the following new section arrangement is completed.}

%\az{The following text moved here from related work:} CountGD-Box and CountVid~\cite{countvid} extend CountGD to output bounding boxes and count objects in videos. By combining text and visual examples, these models achieve superior accuracy over prior approaches.

% In this work, we enhance the counting grammar of CountGD-Box and CountVid. We add the capabilities to specify negative visual and textual examples, automate the selection of visual examples, and allow for them to come from both natural and synthetic external images. These innovations significantly improve the counting accuracy and generality of these models.

To overcome these restrictions, we introduce \methodName, a counting model with novel capabilities in the flexibility of the prompt and its outputs.
Firstly, the prompt is able to specify both \emph{positive} textual and visual examples that describe the object to count as well as any number of \emph{negative} textual and visual examples that describe objects that should {\em not} be counted. The negatives behave as filters, helping the model remove false positives from similar objects as shown in \cref{fig:teaser} (a). We achieve this through a novel use of contrastive training and inference methods that integrate the positives and negatives. Secondly, we automate the identification of visual examples, removing the need for manual annotation. Specifically, bounding boxes output by counting models are cast as {\em `pseudo-exemplars'}  that are fed back to the model for more accurate inference. We show that the pseudo-exemplars significantly improve text only counting performance in images. An example is shown in \cref{fig:teaser} (b). The pseudo-exemplars can also form {\em dynamic} visual examples for objects that change over time in videos. Thirdly, we allow visual examples to come from \emph{external} sources outside of the input image. The representations of these external examples are extracted separately from the input image. We show that these external examples can come from real as well as synthetic images. These capabilities naturally expand on those introduced in previous counting models such as
CLIP-Count~\cite{Jiang2023CLIPCountTT} that allowed the object to be described with text, CountGD~\cite{countgd}, which enabled the use of both positive text and exemplars to
specify targets, and  CountGD-Box~\cite{countvid} that 
additionally outputted boxes.
% Recent CountGD-Box and CountVid extend CountGD, a counting model that accepts visual and textual prompts, to output bounding boxes and count objects in videos. By combining text and visual examples, these models achieve superior accuracy over prior approaches. We add several capabilities to CountGD-Box.
Given these new capabilities, we also  propose  new ways for LLM controllers to use our model as a vision expert agent for counting tasks. 

We  demonstrate these advances improve accuracy, efficiency, and generalization across multiple  datasets including FSCD-147~\cite{m_Ranjan-etal-CVPR21, c_detr}, Blood Cell Detection~\cite{blood_cell_detection}, ShanghaiTech~\cite{Zhang_2016_CVPR}, VideoCount (Crystals)~\cite{countvid}, OmniCount (Fruits)~\cite{omnicount}, PrACo~\cite{mind_the_prompt}, and PairTally~\cite{nguyen2025pairtally}.
All code will be released.

\section{Related Work}
\label{sec:rel_work}
%Expansion of counting grammar

%\az{May need to shorten this section -- in terms of how our work extends prior research.}
% The generality of object counting models has grown over the past few years. In particular, the language for how the object is specified has become richer. In turn, this new versatility has led to significant improvements in counting accuracy and applicability. In this work, we expand on this counting language. 

\noindent\textbf{Closed-world counting. }% No prompts
Object counting methods first developed as class-specific techniques~\cite{Arteta16, 10.1007/978-3-031-19821-2_11, mundhenkLargeContextualDataset2016a, Xie16}. These counting models did not accept any form of object specification. As a result, they were `closed world,' only solving the counting problem for one category of object, such as cars~\cite{Kili2021AnAC}, humans~\cite{10.1007/978-3-031-19821-2_11}, and cells~\cite{Flaccavento11}. Later developments allowed the user to tell the counting model \emph{what} to count, enabling it to adapt to many different objects.

\noindent\textbf{Counting with visual examples. }
The first way to specify the object was visually. These counting methods required the user to manually draw bounding boxes, referred to as `visual exemplars,' over a few example instances inside the image. Given these visual examples, the model would count the remaining instances~\cite{Liu22, dave, Pelhan_2024_NeurIPS, c_detr, low_shot}. Because these visual examples could be provided for any object, these `open-world' methods could count \emph{arbitrary} objects.

% While these methods are very accurate, they suffer from several limitations. Firstly, the visual examples are only used to specify the target object to count. This means there is no mechanism to visually specify what \emph{not} to count. This can make distinguishing between similar objects challenging for these methods. Another issue is that these visual examples must be provided for every image, which can be tedious for large datasets. To address these gaps, our work adds the new capability of specifying negative visual examples and allows the visual examples to come from both synthetic and natural \emph{external} images.

\noindent\textbf{Counting with text. } 
Later methods reduced the annotation burden of visual exemplar-based approaches by using text. Methods like CounTX~\cite{AminiNaieni23}, CLIP-Count~\cite{Jiang2023CLIPCountTT}, and VLCounter~\cite{kang2024vlcounter} allowed the user to specify the object with language rather than bounding boxes. Very recently, text-based methods for counting fine-grained categories of objects have also been developed. GroundingREC~\cite{groundingrec} and CAD-GD~\cite{cad_gd} are trained to distinguish between different attributes, like location and color, of objects.  While these approaches are less tedious to use, requiring no manual annotation, they are unable to benefit from the rich and efficient information in visual examples and cannot output bounding boxes.
\begin{comment}
Other methods tried to automate the selection of visual examples with text. Patch-Selection~\cite{Xu2023ZeroshotOC} identifies `good' visual examples given only the class name and feeds these to a visual-exemplar based counting model. CountSE~\cite{countse} automatically selects soft visual examples implicitly using the text input. However, these methods no longer allow the user to specify explicit visual examples when they are available. This makes it difficult to apply them to settings where explicit visual guidance is necessary. For example, in medicine and materials science, the visual data often differs significantly from the training data and matching objects visually is more accurate.
\end{comment}
% \noindent\textbf{Counting with visual examples \& text. }
More recently, models that accept both visual examples and text have been developed. The current state-of-the-art counting model, CountGD~\cite{countgd} allows users to specify the object with text, visual examples, or both.
%\az{the rest of the text that was here has been moved to the intro}

% \noindent\textbf{Prompting with negatives. } Beyond the counting literature, the use of negatives in related fields is more prominent. For detecting in out-of-distribution images, NegPrompt~\cite{neg_prompt} learns negative prompts using positive class labels. Unlike our approach, the negative prompts cannot be provided explicitly at inference and are instead learned implicitly from the positive prompts. 
% In segmentation and tracking, SAM~2~\cite{ravi2024sam2} allows users to specify \emph{negative clicks} identifying regions that should not be segmented at inference. Similarly, in retrieval, users can provide negative prompts as feedback to refine model predictions~\cite{user_feedback_img_retr, neg_prompt_retr}. Negative prompting has also been widely explored in text-to-image generation methods such as Stable Diffusion~\cite{ban2024understanding}; however, despite its success there, understanding explicit negation remains challenging for VLMs like CLIP~\cite{neg_bench, Radford2021LearningTV}.

%In open-vocabulary detection, methods have explored augmenting loss functions with explicit negative class labels to improve generalisation~\cite{retr_aug_open_vocab_det} \az{If this is only for training/the loss function, then it does not seem to be an example of negative prompting. I would omit it. }. 

\section{The \methodName\ Model}
\label{sec:counting_with_pos_and_neg}

\begin{figure*}[h!]
  \centering
  \includegraphics[width=0.9\linewidth]{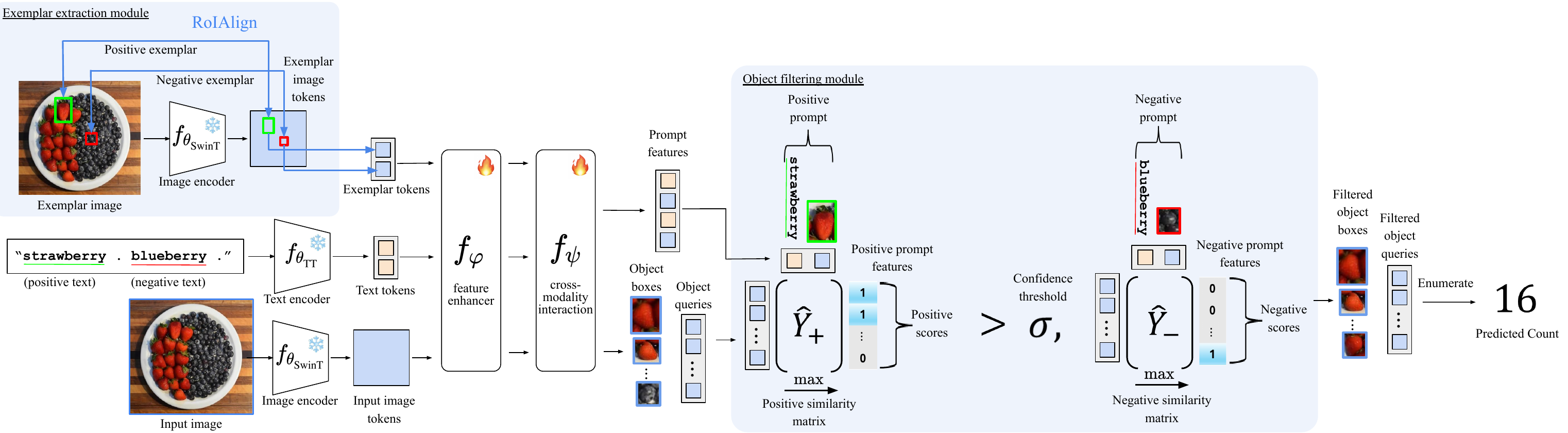}
  \vspace{-2mm}
  \caption{Inference with \methodName. At inference, the object to be counted can be specified by a positive text prompt and any number of positive and negative visual exemplars and text prompts. The model outputs bounding boxes that are enumerated to estimate the count for objects matching the positive prompts. The input image and the image from which the exemplars are obtained (optionally the same as the input image) are passed through the image encoder, $f_{\boldsymbol{\theta_{SwinT}}}$, to obtain image tokens, spatial feature maps. The visual exemplar tokens are cropped out of the exemplar image feature map using RoIAlign in the {\bf Exemplar Extraction Module}. The positive and negative texts are passed through the text encoder, $f_{\boldsymbol{\theta_{TT}}}$, to obtain text tokens. In the feature enhancer, $f_{\boldsymbol{\varphi}}$, the positive visual exemplar and positive text tokens are fused together with self-attention. Separately, the negative visual exemplar and negative text tokens are fused together with self-attention. The fused prompt features then cross-attend to the input image features. Further interaction occurs between the input image features and the prompt features in $f_{\boldsymbol{\psi}}$, which outputs enhanced prompt features and object queries, candidate instances that map to object boxes for all the objects specified by both the positive and negative prompts. The {\bf Object Filtering Module} removes object queries that score below a confidence threshold or are more similar to negative prompts than positive prompts. The remaining object queries are enumerated to estimate the final count. The architecture is built on that of Grounding DINO~\cite{liu2023grounding}.}
  \label{fig:inference_architecture}
\end{figure*}

In this section, we describe \methodName, a counting model that accepts \emph{both} positive and negative prompts to specify the object to count, and the objects {\em not} to count in an image.
A user may input a single positive text prompt and any number of positive visual exemplars together with any number of negative text prompts and negative visual exemplars. The exemplars can be taken from the input image or from a different {\em external} image.
The model is illustrated in \cref{fig:inference_architecture}. 
% \cref{fig:positives_and_negatives} shows an example.
In the following, we first describe the architecture and inference, and then describe the objective function and training. Further details are given in the appendix.

%\az{The following should give more details: add a figure to show the training, after describing the method, add a short paragraph discussing the choices for how negatives could be implemented, e.g.\ self-attention, cross attention and to what.}

\subsection{Architecture}
 
To enable the specification of both positive and negative prompts, we  extend the architecture of CountGD-Box~\cite{countvid}, which in turn, is an extension of Grounding DINO~\cite{liu2023grounding}.
CountGD-Box is a transformer based counting model that accepts {\em only positive} visual exemplars and text, and outputs bounding boxes that are enumerated to estimate the count. 
% Current counting models accept visual examples and text that describe the object that \emph{should be counted}. We name these visual and textual prompts \emph{positive} visual exemplars and text. However, We hypothesize that counting models could also benefit from knowing what \emph{should not be counted}. Therefore, we add \emph{negative} visual exemplars and text that communicate this.

\begin{figure}[h!]
  \centering
  \includegraphics[width=0.8\linewidth]{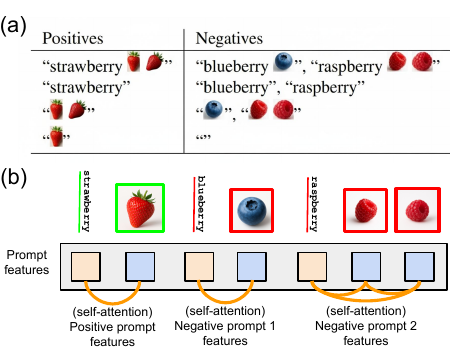}
  \vspace{-3mm}
  \caption{(a) Examples of positive and negative prompts. Any combination is valid. (b) Self-attention between prompt features. In the Feature Enhancer, corresponding visual exemplar and text features self-attend to each other but not to other visual exemplar and text features. Negative prompts do not attend to each other if they describe different classes.}
  \label{fig:prompt_examples_and_self_attention}
\end{figure}

The positive text prompt is denoted as $t^{+}$, and the set of positive visual exemplars is denoted as $\mathbf{B^{+}}$. The negative text prompts and corresponding negative visual exemplars are denoted as a set of pairs $\{(\mathbf{B^{-}_{i}}, t^{-}_{i})\}_{i=0}^{i=N}$, where $\mathbf{B^{-}_{i}}$ contains visual exemplars of the class specified by $t^{-}_{i}$.
% Note that $N$ may equal 0, and pairs of negatives may have no visual exemplars or empty texts. However, if $t^{-}_{i}$ is empty, then $\mathbf{B^{-}_{i}}$ must not be and vice versa. Similarly, if $t^{+}$ is empty, then $\mathbf{B^{+}}$ must not be and vice versa. 
For example, setting $t^{+}=$``strawberry" and $t^{-}_{1}=$``blueberry", $t^{-}_{2}=$``raspberry'', any combination of the positive prompts and the negative prompts in \cref{fig:prompt_examples_and_self_attention} (a) would constitute a valid sentence.

\noindent\textbf{Image Encoder ($f_{\boldsymbol{\theta_{\text{SwinT}}}}$).} The image encoder $f_{\boldsymbol{\theta_{\text{SwinT}}}}$ is a Swin Transformer~\cite{Liu2021SwinTH} that encodes three types of inputs: the input image $X_{input}$ and the positive and negative visual exemplar images $\mathbf{X^{+}}, \mathbf{X^{-}_{i}}$. The same weights are reused for all the inputs. $f_{\boldsymbol{\theta_{\text{SwinT}}}}$ produces spatial feature maps at different scales that are upsampled, concatenated, and projected to 256 dimensions with $1\times1$ convolutions to produce image tokens, feature vectors of length 256 corresponding to the image patches. The input image tokens, $f_{\boldsymbol{\theta_{\text{SwinT}}}}(X_{input})$, are directly input into the Feature Enhancer, $f_{\boldsymbol{\varphi}}$. For the visual exemplars, in the Exemplar Extraction Module, region-of-interest pooling, RoIAlign~\cite{he2017maskrcnn}, is applied to the positive exemplar image tokens, $f_{\boldsymbol{\theta_{\text{SwinT}}}}(X^{+})$, and each of the negative exemplar image tokens, $f_{\boldsymbol{\theta_{\text{SwinT}}}}(X^{-}_{i})$, with the pixel coordinates specified by the corresponding visual exemplars. This process produces one 256-dimensional feature vector for each exemplar. These exemplar tokens are then input into the Feature Enhancer.

\noindent\textbf{Visual exemplars.} CountGD-Box~\cite{countvid} imposes that the input image, $X_{input}$, is the same as the exemplar image, $X^{+}$. 
% While visual exemplars are important since text alone may not be the most effective way to represent the object, manually drawing a box for every input image can also be tedious.
We extend CountGD-Box to accept both positive and negative visual examples from external images. These can be natural or synthetic images that are different from the image used for counting. More formally, we allow $X_{input}$ to be the same as \emph{or different from} $X^{+}$ and each of the $N$ $X^{-}_{i}$'s. To achieve this, as depicted in \cref{fig:inference_architecture}, the input image and the exemplar image are processed in separate streams rather than one unified stream by the same image encoder. Exemplars coming from images different from the input image are  referred to as  {\bf external visual exemplars}. 
The external exemplars allow the user to visually describe the object but only need to be annotated once. After this initial annotation, they can be applied to any number of images without any further annotation. Previously, when using CountGD-Box, a user would need to annotate every image in a dataset to count a particular object in each image. With \methodName, the user only needs to annotate a single image and apply that visual example to the remaining images, significantly reducing the annotation burden.

\noindent\textbf{Text Encoder ($f_{\boldsymbol{\theta_{\text{TT}}}}$).} The text encoder, $f_{\boldsymbol{\theta_{\text{TT}}}}$, is the BERT-base~\cite{bert} text transformer pretrained on detection and phrase grounding data with the image encoder, $f_{\boldsymbol{\theta_{\text{SwinT}}}}$. The text is input in the format ``$t^{+}\text{ }.\text{ }t^{-}_{1}\text{ }.\text{ }t^{-}_{2}\text{ }.\text{ }\cdots\text{ }.\text{ }t^{-}_{N}\text{ }.$" For example, in \cref{fig:inference_architecture}, the input text is ``strawberry . blueberry ." with $t^{+}=$``strawberry" and $t^{-}_{1}=$``blueberry". The text encoder outputs text tokens, 256-dimensional vectors corresponding to the positive and negative text inputs. The text tokens are then input into the Feature Enhancer.

\noindent\textbf{Feature Enhancer ($f_{\boldsymbol{\varphi}}$).} In the Feature Enhancer, $f_{\boldsymbol{\varphi}}$, there is a choice for how the self-attention is applied between the text and visual exemplar prompts. For example, positive and negative prompts could attend to each other. We make the choice to apply self-attention only between text and visual exemplars corresponding to each other, but not between prompts that correspond to different classes. This allows the model to learn to effectively fuse information about the same object~\cite{countgd} while preventing unrelated concepts from influencing each other~\cite{liu2023grounding}. This means prompts from different negative classes do not attend to each other as they may be unrelated. \cref{fig:prompt_examples_and_self_attention} (b) illustrates our self-attention strategy with an example. We ablate other options in the appendix.
The Feature Enhancer is composed of 6 blocks that first fuse the visual exemplar tokens with the text tokens through this self-attention, and then fuse the combined prompt features with the input image patch tokens with cross-attention. $f_{\boldsymbol{\varphi}}$ outputs enhanced input image tokens and prompt features.

\noindent\textbf{Cross-Modality Interaction ($f_{\boldsymbol{\psi}}$).} Further interaction between the input image tokens and the prompt features occurs in $f_{\boldsymbol{\psi}}$. Following \cite{countgd}, the top $k=900$ enhanced input image tokens that are most similar to the enhanced prompt features are first selected. The 900 output tokens from this operation are denoted as `cross-modality queries.' They are passed through a decoder composed of 6 blocks where first the queries self-attend to each other, then cross-attend to the enhanced input image tokens, and finally cross-attend to the enhanced prompt features. $f_{\boldsymbol{\psi}}$ outputs 900 object queries, feature vectors mapping to bounding boxes that localise candidate object instances for \emph{all} objects specified by the prompts. It also carries over the enhanced prompt features from the Feature Enhancer.

\noindent\textbf{Final inference.} The task is now to determine for an object query $q$ whether or not to count it. To this end, we check the following two conditions:\\ $max(Sig(q^{T}P^{+})) > \sigma \text{ \textbf{and} } max(q^{T}P^{+}) > max(q^{T}P^{-})$
\noindent where $P^{+}$ and $P^{-}$ are matrices with the positive and negative prompt features in their columns respectively, and the $max$ is over the prompts, $\sigma$ is a confidence threshold, and $Sig$ is the Sigmoid function.
The first condition checks whether the highest similarity score between the query $q$ and the positive prompts is above the confidence threshold. This is necessary to reject queries that do not map to instances of either the positive or negative prompts. The second condition checks that $q$ is more similar to the positive prompts than to {\em all} the negative prompts. Adding negatives improves the precision of the model, since queries corresponding to negative objects can now be rejected.
% , allowing it to distinguish between objects that have similar properties to the positive class (e.g., similar shape, semantic category, appearance) but are actually of the negative class. 
The object queries that meet these two conditions are enumerated to get the count. Bounding boxes are obtained by passing these queries through an MLP regression head that predicts the normalised box coordinates.

\begin{comment}
The Object Filtering Module, depicted in \cref{fig:inference_architecture}, removes false positives as well as instances of the negative classes. Denote $q_{i}$ as an object query in the form of a row vector, $P^{+}$ as a matrix with the enhanced positive prompt features in its columns, and $P^{-}$ as a matrix with the enhanced negative prompt features in its columns. Let $\sigma$ be a confidence threshold set through a grid search on the validation set. $q_{i}$ will be retained if it meets the condition:

\begin{align*}
max(\hat{y}^{+}_{i}) > \sigma, max(\hat{y}^{-}_{i}) &\\
\hat{y}^{+}_{i} = Sigmoid(q_{i}\times P^{+}) &\\
\hat{y}^{-}_{i} = Sigmoid(q_{i}\times P^{-})
\end{align*}
with $\hat{y}^{+}_{i}$ being the $i^{th}$ row of the positive similarity matrix $\boldsymbol{\hat{Y}_{+}}$, and $\hat{y}^{-}_{i}$ being the $i^{th}$ row of the negative similarity matrix $\boldsymbol{\hat{Y}_{-}}$ in \cref{fig:inference_architecture}.

Intuitively, this means that object queries will be retained if they are similar enough to the positive prompts and are more similar to the positive prompts than to the negative prompts. This improves the precision of the model as it allows it to distinguish between objects that have similar properties to the positive class (e.g., similar shape, semantic category, appearance) but should not be counted. The remaining object queries are enumerated to estimate the count.
\end{comment}

\subsection{Training}

\noindent\textbf{Training objective.} 
To enable the specification of negatives, the model must learn that target object queries should be closer to positive prompt features than to negative ones. More specifically, for a query $q$, positive prompt feature $p^{+}$, and negative prompt feature $p^{-}$, we want $q^{T}p^{+} > q^{T}p^{-}$.

Applying a Sigmoid to both sides of the inequality converts the inner products to probabilities. Introducing the notation $\hat{y}$ for the model predictions, and setting
$\hat{y}^{+} = Sig(q^{T}p^{+})$ and $\hat{y}^{-} = Sig(q^{T}p^{-})$, the inequality clearly holds if we have $\hat{y}^{+}\rightarrow 1$, and $\hat{y}^{-}\rightarrow 0$. This can be learned by applying the Binary Cross Entropy Loss to $\hat{y}^{+}$ with a label of 1 and $\hat{y}^{-}$ with a label of 0. Placing all the queries into the matrix $Q$, and all the prompt features into the matrix $P$, we have $\hat{Y} = Sig(Q^{T}P)$, a matrix that contains these probabilities. The ground truth matrix $Y$ can be constructed such that an entry $y_{i, j} = 1$ if query $i$ corresponds to prompt feature $j$ and 0 otherwise. This objective leads to $\mathcal{L}_{cls} = FocalLoss(\hat{Y}, Y)$, the entrywise Binary Focal Cross Entropy Loss on $\hat{Y}$ and $Y$. 

To train the model to accurately localise object instances, we add three terms from \cite{countvid} to the loss: $\mathcal{L}_{center}$, $\mathcal{L}^{e}_{h,w}$ and $\mathcal{L}^{e}_{GIoU}$. $\mathcal{L}_{center}$ is the sum of the absolute differences between predicted and ground truth object centers. $\mathcal{L}^{e}_{h,w}$ is the sum of the absolute errors of the height and widths and $\mathcal{L}^{e}_{GIoU}$ is the generalized intersection over union between predicted and ground truth boxes. The total loss becomes:\\ $\mathcal{L} = \lambda_{loc}(\mathcal{L}^{e}_{h,w} + \mathcal{L}_{center}) + \lambda_{GIoU}\mathcal{L}^{e}_{GIoU} + \lambda_{cls}\mathcal{L}_{cls}$.

\noindent where $\lambda_{loc}, \lambda_{GIoU}, \lambda_{cls}$ are training hyperparameters selected with a grid search on the validation set.
Note, the `e' superscript indicates exemplars here, since 
bounding boxes are only provided for the exemplars in the FSC-147~\cite{m_Ranjan-etal-CVPR21} training dataset we use. The matching between predicted and ground truth boxes is obtained using the Hungarian Matching Algorithm with the same cost, $\mathcal{C}$, as in \cite{countgd}: $\mathcal{C} = \lambda_{cls}\mathcal{L}_{cls} + \lambda_{loc}\mathcal{L}_{center}$. 

The main difference between our loss and the loss for CountGD-Box~\cite{countvid} is that, unlike in~\cite{countvid}, different object queries within an image may be matched to different classes and prompt features. As a result, our $\mathcal{L}_{cls}$ will push object queries away from visual exemplar and text features they do not correspond to and close to ones they do. This enables the model to separate object queries in embedding space when they correspond to different classes in the input prompt at inference.

\noindent\textbf{Training dataset.} 
For training we require samples where we can specify both positive and negative object categories, so that the model can learn to distinguish between different object categories within an image. These are not available in standard counting datasets (as they usually only have one class to count labeled per image)~\cite{m_Ranjan-etal-CVPR21}.  To address this, we apply the mosaic construction from~\cite{Liu22} to synthesize training images with multiple categories of objects labeled. Example mosaics and further details are in the appendix. 

\section{Automatically Detecting Visual Examples}\label{sec:automatic_detection}
In this section we introduce the concept of {\em pseudo-exemplars} that automate the annotation of the visual exemplars, given a text prompt.
% This capability is used in 
% Sometimes, it can be tedious to manually annotate visual exemplars in every image. For instance, in datasets with thousands of images, it would be necessary to manually annotate thousands of visual examples. To remove this requirement, we introduce the concept of {\em pseudo-exemplars} that automate the annotation of the visual exemplars, given a text prompt.

%\az{Flesh this out: needs a figure to illustrate method, and wrapping this as part of an iterative agent. Maybe include here the video example and the perspective example.}
%\input{figures/automatic-detection}

% pseudo-exemplars for images
Given only text prompts, the model outputs bounding boxes for all the specified objects. We cast $N$ of the top scoring output boxes as `pseudo-exemplars' and feed them back to the model with the text for a second forward pass, producing the final output for counting. In this way, our model can benefit from the rich visual information in the exemplars without requiring any manual annotation from the user. An example is shown in \cref{fig:teaser} (b).

$N$, the number of pseudo-exemplars selected, can be chosen based on the problem setting. For example, in the standard training set for counting, there are three visual exemplars annotated per image, so $N=3$ is a natural choice as the model has been trained for this setting. In applications with fewer than 3 instances, any $N < 3$ may be chosen.  

When both positive and negative text are available, \methodName\ can output both positive and negative pseudo-exemplars. Inputting the positive and negative text as described in \cref{sec:counting_with_pos_and_neg} means the Cross-Modality Interaction step produces object queries for objects either matching the positive or negative prompt. Positive pseudo-exemplars are obtained by selecting the top $N$ boxes from the highest scoring object queries that meet the confidence threshold and are more similar to the positive prompt than to the negative prompt. Similarly, negative pseudo-exemplars are obtained by selecting the top $N$ boxes from the highest scoring object queries that meet the confidence threshold and are more similar to the negative prompt than to the positive prompt. Once the positive and negative pseudo-exemplars have been selected, they can be fed back to the model as though they were manually annotated.

\noindent\textbf{Related methods.} Here we briefly discuss related but different prior methods that also automate exemplar selection. Patch-Selection~\cite{Xu2023ZeroshotOC} is a framework for text-only counting that first selects `good' patches as visual exemplars using a separate error prediction model and then feeds them to an exemplar-only counting model. Unlike this framework, \methodName\ is a unified approach, where the counting model both produces and leverages the selected visual exemplars. \methodName\ also achieves superior accuracy as the counting model benefits from \emph{both} the visual exemplars and the text input. CountSE~\cite{countse} is a text-only counting method that generates `soft' visual exemplars. Unlike \methodName, these soft exemplars are {\em implicit} and do not necessarily encapsulate a single object (see Fig. 3 of \cite{countse}), and the model cannot benefit from real visual exemplars when they are available.
% CountSE also does not output bounding boxes and often lags behind in performance.

\section{\methodName\ as an Expert Agent}\label{sec:expert}
\begin{figure}[h!]
  \centering
  \includegraphics[width=0.85\linewidth]{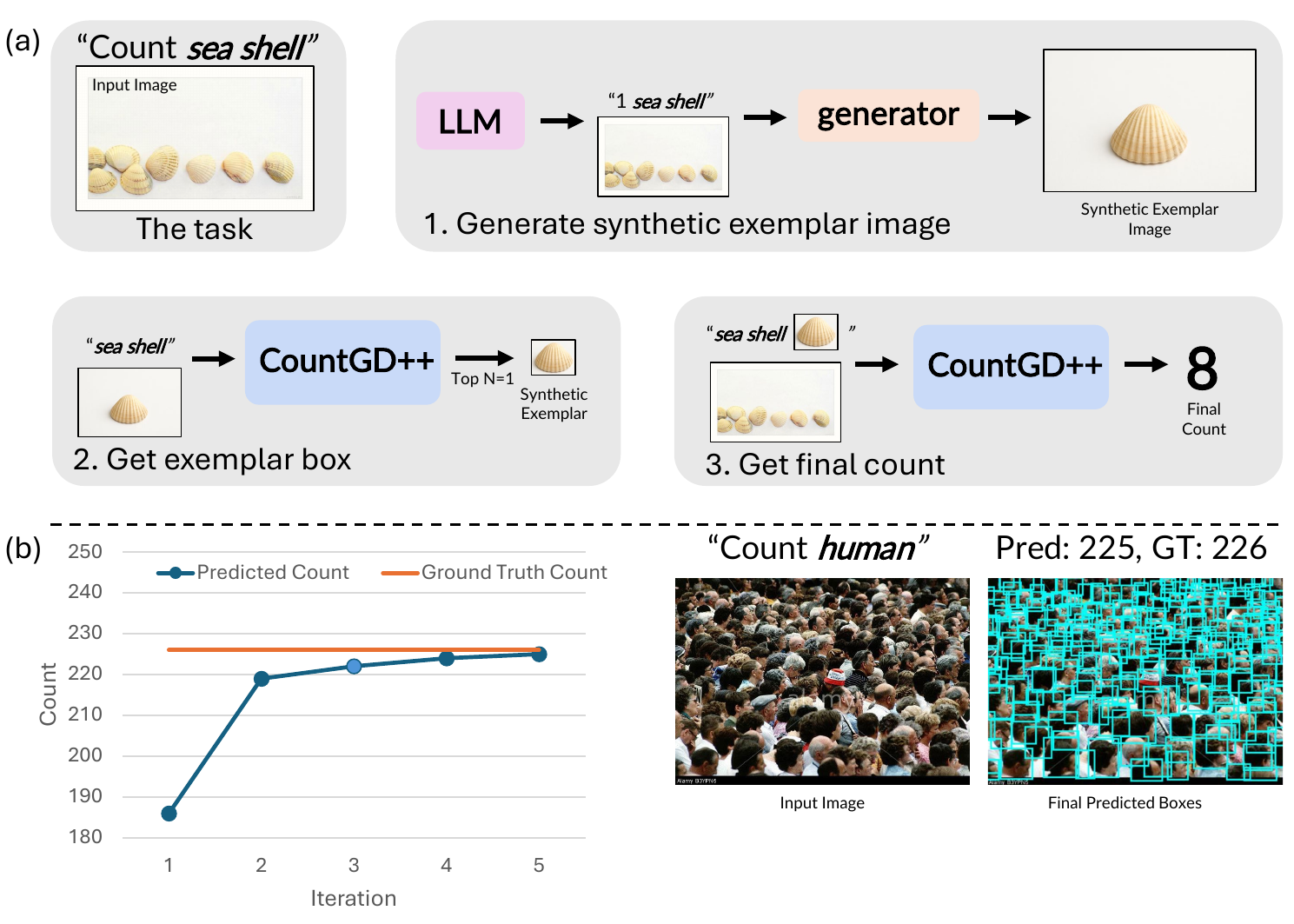}
  \vspace{-3mm}
  \caption{(a) Pipeline for generating and counting with synthetic exemplars using an LLM and \methodName. (b) Example of iteratively improving the count with pseudo-exemplars.}
  \label{fig:expert_images}
\end{figure}

\begin{figure*}[t!]
  \centering
  \includegraphics[width=0.9\linewidth]{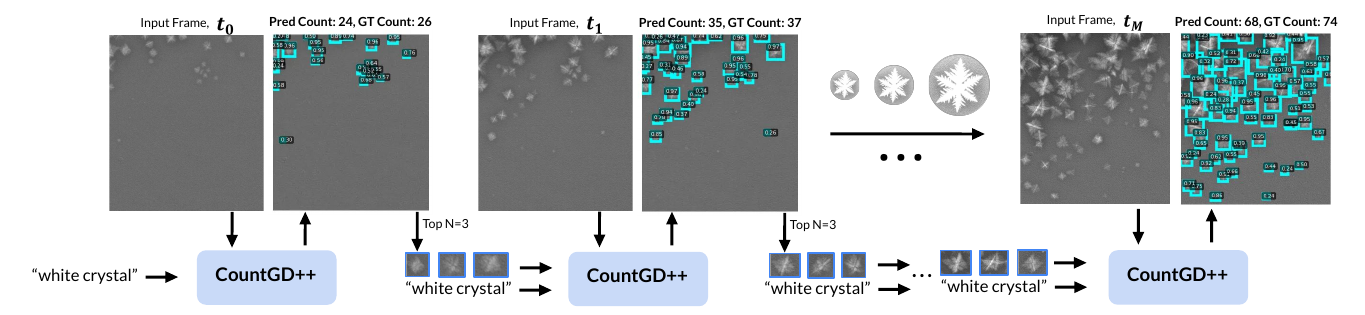}
  \vspace{-3mm}
  \caption{Counting growing and deforming crystals in x-ray videos with pseudo-exemplars. \methodName\ is applied to each frame. In the initial frame, only text is provided. For subsequent frames, the top 3 highest scoring boxes from the previous frame are selected as pseudo-exemplars and input to \methodName\ together with text to predict boxes for the current frame.}
  \label{fig:automatic_detection_videos}
\end{figure*}

Recent work such as ViperGPT~\cite{vipergpt}, HuggingGPT~\cite{hugginggpt}, CodeVQA~\cite{CodeVQA}, Sketchpad~\cite{sketchpad}, AoTD~\cite{aotd}, and VideoAgent~\cite{videoagent} demonstrates that large language models (LLMs) can act as high-level planners that call specialized vision models as expert tools to solve complex multi-modal tasks. With new capabilities like positive and negative prompts, accepting external images, and automatically detecting visual exemplars, \methodName\ can be used as an expert agent by LLMs to improve the performance of counting for images and videos. We describe three examples of how this is done (illustrated in figures~\ref{fig:expert_images} and~\ref{fig:automatic_detection_videos}).
%\az{The rest of this section is too hesitant -- with `would' and `could' everywhere. It reads like a grant proposal for future work. If possible, we should change the tone to be more like: with these new capabilities, CountGD++ can be used as an expert agent by LLMs to improve the performance of counting for images and videos. Each example then shows how it is done (not how it {\em could} be done).}

\noindent\textbf{Example 1: Synthetic exemplars.} In the first setting, we consider an LLM Controller tasked to count objects in an image using only text prompts. Instead of only providing the text to \methodName\ to get the count, the LLM also obtains \emph{external synthetic exemplars} to improve the counting accuracy further. It first calls an API function for image generation, such as one leveraging diffusion models~\cite{hugginggpt}, to synthesize an image with one instance of the object conditioned on the input image and text prompt. If a negative text prompt is provided, the LLM also requests for an image with one instance of the object to not count to be generated. To extract the exemplar from the synthetic image, the controller needs a bounding box of the one instance. This is obtained by applying \methodName\ to the synthetic image with the text and extracting the top scoring box. The synthetic exemplar, text prompt, and input image are fed to a counting API function that leverages \methodName\ as a \emph{counting expert agent} to return the final count to the user. An example is shown in \cref{fig:expert_images} (a).
\begin{comment}
The Controller then passes the synthetic images, input image, and text prompts to a counting API function that leverages \methodName\ as a \emph{counting expert agent} to return the final count to the user. The counting function applies \methodName\ to the synthetic images with the text prompt and extracts the top scoring box from each image as the \emph{synthetic exemplar}. The synthetic exemplars, text prompt, and input image are then be passed to \methodName\ to obtain the final count. This application is possible since \methodName\ accepts external visual examples. 
\end{comment}

\noindent\textbf{Example 2: Iterative agent for images.} Here we consider how an LLM Controller uses the new concept of pseudo-exemplars to iteratively refine the count. Given only a text prompt, the LLM calls a counting API function that leverages \methodName\ to output counts, bounding boxes, and confidence scores for all the objects specified by the prompts. The Controller picks the top $N$ boxes as pseudo-exemplars to pass back to the counting function for another iteration until the count has converged. We refer to this LLM Controller as an \emph{Iterative Agent for Images}. An example of the count improving after each iteration is shown in \cref{fig:expert_images} (b). This operation is similar to `query expansion' in retrieval, where high ranked retrievals are used as the query for another search~\cite{Buckley95,Chum07b}.

\noindent\textbf{Example 3: Iterative agent for videos.} We also consider an LLM tasked to count objects in videos. In video counting, the aim is to count all the unique objects specified by the prompt~\cite{countvid}. The prompts may include any combination of positive text and visual exemplars from a single frame. These prompts are then applied to every frame in the video.

While this works well when objects do not change appearance over time, for objects that evolve, this approach is not ideal. This is because the visual exemplars from the first frame may not look visually similar to the same objects in future frames. 
To address this issue, the LLM calls a video counting function that applies `pseudo-exemplars' to videos. The video counting function calls the image counting model on each frame, takes the top $N$ most confident boxes as pseudo-exemplars from the prediction on the current frame and passes them together with the text to the image counting model to process the next frame. This allows the visual exemplars to automatically update themselves and evolve over time as the objects do, leading to significantly more accurate video counting for deforming objects. A schematic for this approach and an example application are shown in \cref{fig:automatic_detection_videos}.
\section{Experiments}
\label{sec:experiments}
\methodName\ is trained on the FSC-147~\cite{m_Ranjan-etal-CVPR21} object counting training set, and then evaluated on the FSCD-147~\cite{c_detr} test set, and seven other benchmark datasets (without any fine-tuning). We first describe the datasets, and then discuss the performance.
% TO DO Add GroundingREC results as a baseline on other datasets to show 'lags behind' due to no visual exemplars, say cannot evaluate CAD-GD since pretrained checkpoints not available, consider retraining from scratch
\subsection{Datasets \& Metrics}
\noindent\textbf{Datasets.} FSC-147~\cite{m_Ranjan-etal-CVPR21} is the standard open-world counting dataset covering 147 classes and 6135 images with 7-3731 objects per image. FSCD-147~\cite{c_detr} adds bounding boxes to the validation and test sets of FSC-147. PrACo~\cite{mind_the_prompt} is a counting benchmark constructed from images in FSC-147. It introduces the `Negative Label Test' to evaluate counting models when the target object is not in the image, and the `Mosaic Test' to evaluate counting models in the multi-class setting. We also test our model without any fine-tuning on the ShanghaiTech crowd counting dataset~\cite{Zhang_2016_CVPR} (498 images with 9-2256 humans per image), Blood Cell Detection dataset for counting red and white blood cells~\cite{blood_cell_detection} (100 microscopic images with 11-33 cells from a peripheral blood smear), the Fruits subset of OmniCount-191~\cite{omnicount} (303 images with 3-6 instances of 8 different fruits per image), the CARPK~\cite{Hsieh2017DroneBasedOC} test set (459 aerial drone images of cars with 2-188 cars per image), and the PairTally~\cite{nguyen2025pairtally} Benchmark for fine-grained object counting (681 dense images (e.g., 200+ instances) of mixed objects with subtle differences). We additionally test adding pseudo-exemplars to CountVid~\cite{countvid} on the Science-Count (Crystals) benchmark of VideoCount containing 7 videos with 10-154 dynamic crystals rapidly forming in x-ray videos. Results on PairTally and CARPK and further details on the choice of datasets are in the appendix. 

\noindent\textbf{Metrics.}
% counting MAE, RMSE, Detection AP, AP50 for images
To evaluate counting accuracy in images, we use the image-based Mean Absolute Error (MAE) and Root Mean Squared Error (RMSE) from~\cite{countvid, AminiNaieni23, Liu22, countgd}. To evaluate detection accuracy in images, following~\cite{countvid, c_detr, Pelhan_2024_NeurIPS, dave}, the mean average precision over thresholds 0.5 to 0.95 (AP) and the average precision at the IoU threshold of 0.5 (AP50) are used. The Normalized Mean of Negative predictions (NMN), Positive Class Count Nearness (PCCN), Counting Precision (CntPr), and Counting Recall (CntR) introduced in \cite{mind_the_prompt} are also reported on the PrACo Benchmark. For counting in videos, we use the video-based MAE and RMSE defined 
in~\cite{countvid}.

% PrACo metrics
% Counting MAE and RMSE for videos

\subsection{Implementation}
% Need to say we use one pretrained checkpoint for everything and the same hyperparameters and confidence threshold
The coefficients on the loss terms $\lambda_{loc}$, $\lambda_{GIoU}$, and $\lambda_{cls}$ are set to 5, 2, 2 respectively. These hyperparameters are borrowed directly from CountGD-Box~\cite{countvid} with no further tuning. \methodName\ is trained on FSC-147~\cite{m_Ranjan-etal-CVPR21} with 1000 of our synthetic mosaic images added. No fine-tuning is done on any of the other datasets. We use the same confidence threshold $\sigma=0.23$, borrowed from CountGD-Box~\cite{countvid}, across all datasets without further optimization. For FSCD-147, an adaptive cropping scheme that outputs bounding boxes is applied for dense scenes. Further training and inference details are in the appendix.

\subsection{Results}
\begin{figure*}[h!]
  \centering
  \includegraphics[width=0.9\linewidth]{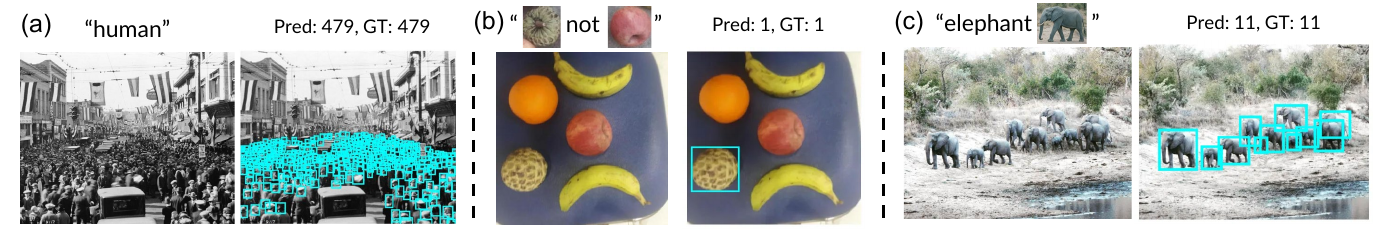}
  \vspace{-3mm}
  \caption{Counting results on images from the test sets. (a) {\bf ShanghaiTech}: positive text and pseudo-exemplars are used to count in dense crowds; (b) {\bf OmniCount (Fruits)}: positive and negative external exemplars are used to distinguish between different types of apples; (c) {\bf FSCD-147}: positive text, synthetic, and pseudo-exemplars are used. The synthetic exemplar is at the top of the image in the quotes.
\label{fig:qual_results}}
\end{figure*}
Here we show that our new capabilities result in superior performance over prior approaches to open-world counting across a wide variety of datasets. Qualitative results are shown in \cref{fig:qual_results}. Due to space limitations, we include results for the essential prompt settings here and refer the reader to the appendix for more exhaustive results.

\begin{table}
\begin{center}
{\fontsize{8}{10}\selectfont\begin{NiceTabular}{l|c|c|c|c} 
\hline
     & \multicolumn{4}{c}{FSCD-147 Test Text Only}\\
   Method & \multicolumn{2}{c}{Counting} & \multicolumn{2}{c}{Detection}\\
   %& Setting & \multicolumn{2}{c}{} & \multicolumn{2}{c}{}\\
   & MAE $\downarrow$ & RMSE $\downarrow$ & AP $\uparrow$ & AP50 $\uparrow$\\
   \hline
   \rowcolor{gray!20} $\text{DAVE}_{prm}$~\cite{dave} & 14.90 & 103.42 & \xmark & \xmark\\
   \rowcolor{gray!20} CountGD~\cite{countgd} & 12.98	& 98.35 & \xmark & \xmark\\
   \rowcolor{gray!20} GrREC~\cite{groundingrec} &  10.12 & 107.19 & \xmark & \xmark\\
   \rowcolor{gray!20} CAD-GD~\cite{cad_gd} &  10.35 & 86.88 & \xmark & \xmark\\
   \rowcolor{gray!20} CountSE~\cite{countse} &  \textbf{7.84} & 82.99 & \xmark & \xmark\\
  PseCo~\cite{zhizhong2024point} &  16.58 & 129.77 & 37.91* & 62.45* \\
  $\text{DAVE}_{prm}$~\cite{dave} &  15.52 & 114.10 & 18.50 & 50.24\\
    CGD-B~\cite{countvid} &  15.01 & 118.16 & 30.44 & 61.56\\
    $\text{Ours}_{t}$ &  16.55 & 129.76 & 33.01 & 61.75\\
        $\text{Ours}_{t+p}$ &  10.29 & 33.52 & 37.78 & 68.90\\
    \textbf{$\text{Ours}_{t+p+s}$} &  \textbf{8.39} & \textbf{27.03} & \textbf{38.93} & \textbf{71.35}\\
\hline
\end{NiceTabular}}
\vspace{-1mm}
\caption{\label{tab:fscd147} Results on {\bf FSCD-147} for image counting methods given only positive text input. Results for counting methods that do not output boxes are grayed out. * for PseCo indicates the result was obtained using the published checkpoints and the same bounding boxes for counting and detection. For $\text{Ours}_{t}$, we use no pseudo- or synthetic exemplars. For $\text{Ours}_{t+p}$ we add pseudo-exemplars. $\text{Ours}_{t+p+s}$ also adds synthetic exemplars. The abbreviations are: GroundingREC (GrREC); CountGD-Box (CGD-B).\vspace{-8mm}}
%Results on FSCD-147 Test for image counting methods that output boxes (for Stage~1). `GDINO' is Grounding DINO, `CGD-B' is \countgdbox, and `C-DETR' is Counting-DETR~\cite{10.1007/978-3-031-20044-1_20}. SAM masks are not used in evaluation. `exemp.' means exemplars only, and `both' means both exemplars and text as prompts.}
\end{center}
 
\end{table}

\noindent\textbf{FSCD-147~\cite{m_Ranjan-etal-CVPR21} \& PrACo~\cite{mind_the_prompt}.} In \cref{tab:fscd147} and \cref{tab:praco}, we test \methodName\ on the setting where only text is available. Using only the class name, and employing GPT-5~\cite{gpt5} as the LLM in the pipeline from \cref{sec:expert}, synthetic exemplars are generated for FSCD-147. We test \methodName\ with and without the pseudo- and synthetic exemplars on FSCD-147 when there is only one class labeled per image. 1 synthetic exemplar and text are used in the first forward pass. 3 pseudo-exemplars and text are used in the final forward pass. We use both positive and negative pseudo- and synthetic exemplars when both positive and negative text are available in PrACo.
By using pseudo- and synthetic exemplars, \methodName\ achieves the best counting RMSE and detection accuracy over all text-only methods. CountSE~\cite{countse} has a similar MAE but a much higher RMSE and does not output boxes. Clearly, adding pseudo- and synthetic exemplars improves performance. \cref{tab:praco} demonstrates by leveraging both positive and negative text, \methodName\ outperforms the other models on PrACo.

\begin{table*}
\begin{center}
{\fontsize{8}{10}\selectfont\begin{NiceTabular}{l|l|l|l|l|l|l|c|c|c|c|c|c|c|c} 
% \fontsize{9}{11}\selectfont\begin{NiceTabular}{l|l|l|l|l|l|l|c|c|c|c|c|c|c|c} 
\hline

     & \multicolumn{6}{c}{Prompt} & \multicolumn{4}{c}{Blood Cell Detection} & \multicolumn{4}{c}{OmniCount (Fruits)}\\
    
   %\hline
   Method & \multicolumn{3}{c}{Positives} & \multicolumn{3}{c}{Negatives} & \multicolumn{2}{c}{Counting} & \multicolumn{2}{c}{Detection} & \multicolumn{2}{c}{Counting} & \multicolumn{2}{c}{Detection}\\
   %& Setting & \multicolumn{2}{c}{} & \multicolumn{2}{c}{}\\
   &  $t^{+}$ & $B^{+}_{int}$ & $B^{+}_{ext}$ & $t^{-}$ & $B^{-}_{int}$ & $B^{-}_{ext}$& MAE $\downarrow$ & RMSE $\downarrow$ & AP $\uparrow$ & AP50 $\uparrow$ & MAE $\downarrow$ & RMSE $\downarrow$ & AP $\uparrow$ & AP50 $\uparrow$\\
   \hline
   
  \rowcolor{gray!20} CountGD~\cite{countgd} & \cmark & \cmark & \xmark & \xmark & \xmark & \xmark & 10.99	& 14.64	& \xmark & \xmark & 2.76 & 3.11 & \xmark & \xmark\\
   CGD-B~\cite{countvid} & \cmark & \cmark & \xmark & \xmark & \xmark & \xmark & 11.34 & 15.42 & 0.25 & 0.45 & 2.83 & 3.15 & 0.47 & 0.61\\
   Ours & \cmark & \cmark & \xmark & \xmark & \xmark & \xmark & 11.56 & 15.69 & 0.27 & 0.47 & 2.41 & 2.92 & 0.54 & 0.72\\
   Ours & \cmark & \xmark & \cmark & \xmark & \xmark & \xmark & 11.62 & 15.84 & 0.33 & 0.53 & 2.38 & 2.86 & 0.55 & 0.73\\
   \textbf{Ours} & \cmark & \cmark & \xmark & \cmark & \cmark & \xmark & 1.73 & 3.06 & 0.46 & 0.71 & \textbf{0.41} & \textbf{1.51} & \textbf{0.62} & \textbf{0.83} \\
   \textbf{Ours} & \cmark & \xmark & \cmark & \cmark & \xmark & \cmark & \textbf{1.52} & \textbf{2.42} & \textbf{0.54} & \textbf{0.80} & 0.49 & 1.63 & 0.60 & 0.80\\
   \hline

\end{NiceTabular}}
\vspace{-2mm}
\caption{\label{tab:blood_cell_detection_and_omnicount}Results on the {\bf Blood Cell Detection}~\cite{blood_cell_detection} and the {\bf OmniCount (Fruits)}~\cite{omnicount} test sets. The symbols for provided prompts are: positive text ($t^{+}$), 1 positive visual exemplar from inside each image ($B^{+}_{int}$), 1 positive visual exemplar from one external image applied across the dataset ($B^{+}_{ext}$), negative text ($t^{-}$), 1 negative visual exemplar from inside each image ($B^{-}_{int}$), 1 negative visual exemplar from one external image applied across the dataset ($B^{-}_{ext}$). External exemplars may generalize better to other instances in the image when they depict the object more clearly or under more representative conditions than the internal exemplar.\vspace{-6mm}}
\end{center}
 
\end{table*}
\begin{table}
\begin{center}
{\fontsize{8}{10}\selectfont\begin{NiceTabular}{l|c|c|c|c|c|c} 
    \hline
     & \multicolumn{6}{c}{PrACo Test}\\
    Method & \multicolumn{2}{c}{Prompt} & \multicolumn{2}{c}{Negative Test} & \multicolumn{2}{c}{Mosaic Test} \\
    & $t^{+}$ & $t^{-}$ & NMN $\downarrow$ & PCCN $\uparrow$ & CntP $\uparrow$ & CntR $\uparrow$\\
    \hline
    TFPOC~\cite{tfoc} & \cmark & \xmark & 0.75 & 66.04 & 0.69 & 0.85 \\ 
    DAVE~\cite{dave} & \cmark & \xmark & 1.05 & 37.02 & 0.84 & 0.80 \\
    DAVE~\cite{dave} & \cmark & \cmark & 0.08 & 97.62 & 0.84 & 0.80\\
    Ours & \cmark & \xmark & 0.88 & 62.86 &  0.86 & \textbf{0.96} \\
    \textbf{Ours} & \cmark & \cmark & \textbf{0.07} & \textbf{97.99} & \textbf{0.90} & \textbf{0.96} \\
   \hline
   
\hline
\end{NiceTabular}}
\vspace{-2mm}
\caption{\label{tab:praco} Results on the {\bf PrACo}~\cite{mind_the_prompt} benchmark. For Ours we use both positive and negative pseudo- and synthetic exemplars. The symbols for provided prompts are: positive text ($t^{+}$), negative text ($t^{-}$). CntF1 is omitted since it's obtained from CntP and CntR.\vspace{-6mm}}
\end{center}
 
\end{table}
\begin{table}
\begin{center}
{\fontsize{8}{10}\selectfont\begin{NiceTabular}{l|c|c|c|c} 
  \hline
    & \multicolumn{4}{c}{ShanghaiTech Test}\\
   Method & \multicolumn{2}{c}{Part A} & \multicolumn{2}{c}{Part B}\\
   %& Setting & \multicolumn{2}{c}{} & \multicolumn{2}{c}{}\\
   & MAE $\downarrow$ & RMSE $\downarrow$ & MAE $\downarrow$ & RMSE $\downarrow$ \\
   \hline
  GDINO~\cite{liu2023grounding} & 394.9 & 537.5 &  58.3 & 99.3 \\
  OWLv2~\cite{owlv2} & 420.2 & 553.3 & 81.5 & 126.5\\
  CLIP-Count~\cite{Jiang2023CLIPCountTT} & 192.6 & 308.4 & 45.7 & 77.4\\
  CGD-B~\cite{countvid} & 132.2 & 253.9 & 32.2 & 57.9\\
    CountSE~\cite{countse} & 129.7 & 258.3 & \xmark & \xmark\\
  \textbf{Ours} & \textbf{116.0} & \textbf{234.0} & \textbf{28.0} & \textbf{50.0}\\
\hline
\end{NiceTabular}}
\vspace{-2mm}
\caption{\label{tab:shanghai_tech} Counting results on Parts A and B of the {\bf ShanghaiTech}~\cite{Zhang_2016_CVPR} crowd counting dataset given only positive text input. Part A contains fewer images and more crowded scenes than Part B. For Ours we use pseudo-exemplars. Results for CountSE~\cite{countse} on Part B are not available. The abbreviations are: CountGD-Box (CGD-B); Grounding DINO (GDINO).\vspace{-6mm}}
\end{center}
\end{table}
\begin{table}
\begin{center}
{\fontsize{8}{10}\selectfont\begin{NiceTabular}{l|c|c|c|c} 
   \hline
    & \multicolumn{2}{c}{} & \multicolumn{2}{c}{VideoCount (Crystals)}\\
    Method & \multicolumn{2}{c}{Prompt} & \multirow{2}{*}{MAE $\downarrow$} & \multirow{2}{*}{RMSE $\downarrow$}\\
    & $t^{+}$ & $B^{+}_{int}$ & &\\
   \hline
 CountVid~\cite{countvid} & \cmark & \cmark & 12 & 13.5\\
CountVid~\cite{countvid} & \cmark & \xmark & 69.1 & 86\\
\textbf{Ours} & \cmark & \xmark & \textbf{10} & \textbf{12.3}\\
\hline
\end{NiceTabular}}

\caption{\label{tab:crystals} Video counting results on the {\bf Science-Count (Crystals)} dataset. For Ours we use pseudo-exemplars inside each video frame and across the video as shown in \cref{fig:automatic_detection_videos}. The symbols for provided prompts are: positive text ($t^{+}$), the 3-8 manually annotated positive visual exemplars from CountVid ($B^{+}_{int}$).\vspace{-6mm}}
\end{center}
\end{table}
\noindent\textbf{Blood Cell Detection~\cite{blood_cell_detection} \& OmniCount (Fruits)~\cite{omnicount}.} In \cref{tab:blood_cell_detection_and_omnicount}, we test \methodName\ on the multi-class images in Blood Cell Detection and OmniCount (Fruits) given positive text and exemplars, and negative text and exemplars. Exemplars are either \emph{internal}, coming from inside each image, or \emph{external}, coming from one image applied across the dataset. The Positive text is the class name and the negative texts are the names of the other classes in the image. Only one exemplar is provided for each class. The baselines are CountGD~\cite{countgd} and CountGD-Box~\cite{countvid}, since these are the only other two models that accept both text and exemplars, and they are both accurate at counting.
Given positive and negative text, and positive and negative internal or external exemplars, \methodName\ significantly outperforms CountGD and CountGD-Box at counting and detection, sometimes even reducing the MAE by an order of magnitude. This shows by leveraging negative examples, which only it can do, \methodName\ achieves greater accuracy as it is better able to distinguish between similar objects like red and white blood cells. These results also show \methodName\ generalises to external visual exemplars. Given only positive text and exemplars, \methodName\ generally matches the counting performance of CountGD and CountGD-Box while outperforming CountGD-Box at detection. 

\noindent\textbf{ShanghaiTech~\cite{Zhang_2016_CVPR}.} In \cref{tab:shanghai_tech}, we test \methodName\ when only the positive text `human' is available on the ShanghaiTech crowd counting dataset. In the first forward pass, text is input to \methodName. In the final forward pass, text and 3 pseudo-exemplars are used.
\methodName\ achieves state-of-the-art counting performance on both parts of ShanghaiTech, reducing the MAE of CountSE~\cite{countse} by over 10\% and RMSE by 9\% on Part~A. % Results for CountSE are not provided for Part B, and the model is not open source yet.

\noindent\textbf{VideoCount (Crystals)~\cite{countvid}.} In \cref{tab:crystals}, we test the effect of applying pseudo-exemplars to counting evolving objects in videos. We use the published SAM~2~\cite{ravi2024sam2} and CountGD-Box checkpoints from CountVid. For each frame, two forward passes are conducted. In the first forward pass per frame, the text `white crystal' and 10 pseudo-exemplars are provided from the prior frame (with the exception of the initial frame, where only text is used). In the second forward pass per frame, 10 pseudo-exemplars from the current frame and the text are input to the model. The top 10 most confident boxes are then passed as pseudo-exemplars for the next frame. A qualitative example and schematic of this approach with 3 pseudo-exemplars are illustrated in \cref{fig:automatic_detection_videos}.
Pseudo-exemplars significantly improve the performance over CountVid~\cite{countvid}. In the text-only setting, the MAE and RMSE are divided by a factor of about 7. The significant improvement is because the model is now able to match the crystals visually. Our approach remarkably also beats the manually annotated exemplars. This is for two reasons. The first reason is that pseudo-exemplars evolve over time as the crystals do, while the manually annotated exemplars are only from the first frame. The second reason is that we use up to 10 pseudo-exemplars at each frame, which is more than the 3-8 provided to CountVid. Since our approach is automatic, it can provide more visual examples without any additional effort from the user.

\section{Conclusion}
\label{sec:conclusion}
We propose \methodName, a model that introduces the new capabilities of specifying both positive and negative prompts, automating exemplar detection with pseudo-exemplars, and accepting both natural and synthetic external exemplar images. In turn, these new capabilities lead to significant improvements to counting and detection accuracy across several datasets. As new LLMs still have limited counting abilities (see supplementary), we also show how LLMs can use \methodName\ as a Counting Expert Agent for vision applications.

\section*{Acknowledgments}
The authors would like to thank Dr Christian Schroeder de Witt (Oxford Witt Lab, OWL) for his helpful feedback and insights on the paper figures and Gia Khanh Nguyen, Yifeng Huang, and Professor Minh Hoai for their help with the PairTally Benchmark~\cite{nguyen2025pairtally}. This research is funded by an AWS Studentship, the Reuben Foundation, a Qualcomm Innovation Fellowship (mentors: Dr Farhad Zanjani and Dr Davide Abati), the AIMS CDT program at the University of Oxford, EPSRC Programme Grant VisualAI EP/T028572/1, and a Royal Society Research Professorship RSRP$\backslash$R$\backslash$241003.

{
    \small
    \bibliographystyle{ieeenat_fullname}
    \bibliography{longstrings, vgg_local, other}
}
\appendix

\renewcommand{\thefigure}{A.\arabic{figure}} % \thesection instead of A would make it A.1, B.1...
\setcounter{figure}{0} 
\renewcommand{\thetable}{A.\arabic{table}}
\setcounter{table}{0} 

\addcontentsline{toc}{section}{Appendix} % Add the appendix text to the document TOC
\part{Appendix} % Start the appendix part
\parttoc % Insert the appendix TOC

\section{Further Quantitative Results}

\subsection{Results on PairTally~\cite{nguyen2025pairtally}}
\begin{table}[h!]
\begin{center}
{\fontsize{8}{10}\selectfont\begin{NiceTabular}{l|c|c|c|c|c|c} 
   \hline
    & \multicolumn{4}{c}{} & \multicolumn{2}{c}{PairTally}\\
    Method & \multicolumn{4}{c}{Prompt} & \multirow{2}{*}{MAE $\downarrow$} & \multirow{2}{*}{RMSE $\downarrow$}\\
    & $t^{+}$ & $B^{+}_{int}$ &$t^{-}$ & $B^{-}_{int}$ & &\\
   \hline
   Qwen2.5-VL~\cite{Qwen2.5-VL} & \cmark & \xmark & \xmark & \xmark & 59.36 & \xmark\\
   LLaMA-3.2~\cite{llama3} & \cmark & \xmark & \xmark & \xmark & 54.67 & \xmark\\
   GeCo~\cite{Pelhan_2024_NeurIPS} & \xmark & \cmark & \xmark & \xmark & 50.24 & \xmark\\
   DAVE~\cite{dave} & \xmark & \cmark & \xmark & \xmark & 47.37 & \xmark\\
   CountGD~\cite{countgd} & \cmark & \cmark & \xmark & \xmark & 46.67 & 70.85\\
   Ours & \cmark & \cmark & \xmark & \xmark & 46.41 & 69.52 \\
   \textbf{Ours} & \cmark & \cmark & \cmark & \cmark & \textbf{35.27} & \textbf{60.85}\\
\hline
\end{NiceTabular}}
\vspace{-2mm}
\caption{\label{tab:pairtally} Results on the {\bf PairTally}~\cite{nguyen2025pairtally} test set. \xmark\ in the RMSE column indicates the RMSE results were not reported in the original PairTally paper. For CountGD, the MAE was reproduced using the published PairTally code, and the RMSE was obtained using the same code. The symbols for provided prompts are: positive text ($t^{+}$), 3 positive visual exemplars from inside each image ($B^{+}_{int}$), negative text ($t^{-}$), 3 negative visual exemplars from inside each image ($B^{-}_{int}$).}
\end{center}
\end{table}

In \cref{tab:pairtally}, we evaluate \methodName\ on the PairTally Benchmark~\cite{nguyen2025pairtally} for fine-grained object counting. \methodName\ achieves a new state-of-the-art MAE and RMSE on this benchmark. Provided with only positive prompts, \methodName\ beats CountGD~\cite{countgd}. Providing both positive and negative prompts improves the counting accuracy further. These results are particularly impressive as PairTally is a very challenging dataset, including high counts of mixed objects with subtle differences in shape, color, and texture. Qualitative results are shown in \cref{fig:qual_results_pairtally_supp}.

\subsection{Results on CARPK Test~\cite{Hsieh2017DroneBasedOC}}
\begin{table}
\begin{center}
{\fontsize{8}{10}\selectfont\begin{NiceTabular}{l|c|c|c|c} 
  \hline
    & \multicolumn{2}{c}{} & \multicolumn{2}{c}{CARPK Test}\\
    Method & \multicolumn{2}{c}{Prompt} & \multirow{2}{*}{MAE $\downarrow$} & \multirow{2}{*}{RMSE $\downarrow$}\\
    & $t^{+}$ & $B^{+}_{int}$ & &\\
   \hline
CLIP-Count~\cite{Jiang2023CLIPCountTT} & \cmark & \xmark & 11.96 & 16.61\\
CounTX~\cite{AminiNaieni23} & \cmark & \xmark & 8.13 & 10.87\\
VLCounter~\cite{kang2024vlcounter} & \cmark & \xmark & 6.46 & 8.68\\
CountGD~\cite{countgd} & \cmark & \xmark & 3.83 & 5.41\\
CountGD~\cite{countgd} & \cmark & \cmark & 3.68 & 5.17\\
CountSE~\cite{countse} & \cmark & \xmark & 2.79 & 4.20\\
\textbf{Ours} & \cmark & \xmark & \textbf{2.48} & \textbf{3.74}\\
\hline
\end{NiceTabular}}
\vspace{-2mm}
\caption{\label{tab:carpk} Counting results on the {\bf CARPK}~\cite{Hsieh2017DroneBasedOC} car counting dataset given only positive text input. For Ours we use 3 pseudo-exemplars, detected with text only. The symbols for provided prompts are: positive text ($t^{+}$), 3 positive visual exemplars manually annotated inside each image ($B^{+}_{int}$).\vspace{-6mm}}
\end{center}
\end{table}

In \cref{tab:carpk}, we evaluate \methodName\ on the CARPK Test Set~\cite{Hsieh2017DroneBasedOC} for car counting from drone imagery. We only use the positive text ``car'' as the prompt to \methodName.  Pseudo-exemplars are detected by \methodName\ on the first forward pass given only this text prompt, and then fed back to the model together with the positive text for the final prediction. \methodName\ achieves a new state-of-the-art MAE and RMSE on CARPK, beating both methods that accept text only and methods that accept both visual exemplars and text.

\subsection{Results in More Prompt Settings}

While in the main paper we only had room for presenting the results on FSCD-147~\cite{m_Ranjan-etal-CVPR21} given text only, in \cref{tab:fscd147_full} we also include results given the three internal (provided) exemplars only, or both the text and the internal (provided) exemplars. In the text-only prompt setting, our approach also uses the synthetic and pseudo-exemplars, since these are generated automatically given text only. In all settings, the adaptive cropping procedure described in \cref{subsec:inference} is applied. Furthermore, in the settings where internal exemplars are provided, the SAM Test-time normalization introduced in \cite{countgd} is applied to help avoid double counting. Note, the AP and AP50 are calculated based on the boxes after adaptive cropping but not test-time normalization. This is because the test-time normalization does not improve the bounding boxes, only the predicted count. Please see the caption of \cref{tab:fscd147_full} for further details.

In the text-only setting, \methodName\ is the superior method, achieving the best detection accuracy and the lowest counting RMSE among both methods that can and cannot output bounding boxes. Given exemplars only, \methodName\ achieves the lowest counting MAE for all methods and the lowest counting RMSE among methods that output boxes. \methodName's detection accuracy is significantly better than CountGD-Box's~\cite{countvid}, and its counting accuracy is much better than CountGD's~\cite{countgd}. In the multi-modal setting, when both text and internal exemplars are provided, \methodName\ achieves competitive counting accuracy with CountGD, significantly better counting accuracy than CountGD-Box, and significantly better detection accuracy over CountGD-Box.

\begin{table}
\begin{center}
{\fontsize{9}{11}\selectfont\begin{NiceTabular}{l|c|c|c|c|c} 

    \multicolumn{2}{c}{} & \multicolumn{4}{c}{FSCD-147 Test}\\
   \hline
   \multirow{3}{*}{~~~~~~~Method} & \multirow{3}{*}{Prompt} & \multicolumn{2}{c}{Counting} & \multicolumn{2}{c}{Detection}\\
   %& Setting & \multicolumn{2}{c}{} & \multicolumn{2}{c}{}\\
   & & MAE & RMSE & AP & AP50 \\
   & & $\downarrow$ & $\downarrow$ & $\uparrow$ & $\uparrow$\\
   \hline
   \multicolumn{6}{c}{Text Only}\\
   \rowcolor{gray!20} $\text{DAVE}_{prm}$ & $t^{+}$ & 14.90 & 103.42 & \xmark & \xmark\\
   \rowcolor{gray!20} CountGD & $t^{+}$ & 12.98	& 98.35 & \xmark & \xmark\\
   \rowcolor{gray!20} T2ICount & $t^{+}$ & 11.76 & 97.86 & \xmark & \xmark\\
   \rowcolor{gray!20} GrREC & $t^{+}$ & 10.12 & 107.19 & \xmark & \xmark\\
   \rowcolor{gray!20} CAD-GD & $t^{+}$ & 10.35 & 86.88 & \xmark & \xmark\\
   \rowcolor{gray!20} CountSE & $t^{+}$ & \textbf{7.84} & 82.99 & \xmark & \xmark\\
  GDINO & $t^{+}$ & 54.16 & 157.87 &  11.60 & 17.80 \\
  OWLv2 & $t^{+}$ & 41.83 & 149.82 & 22.84 & 35.76\\
  PSeCo & $t^{+}$ & 16.58 & 129.77 & 37.91* & 62.45* \\
  $\text{DAVE}_{prm}$ & $t^{+}$ & 15.52 & 114.10 & 18.50 & 50.24\\
    CGD-B & $t^{+}$ & 15.01 & 118.16 & 30.44 & 61.56\\
    \textbf{Ours} & $t^{+}$ & 8.39 & \textbf{27.03} & \textbf{38.93} & \textbf{71.35}\\
    \hline
       \multicolumn{6}{c}{Internal (Provided) Exemplars Only}\\
    \rowcolor{gray!20} DAVE & $B^{+}_{int}$ & 8.66 & \textbf{32.36} & \xmark & \xmark\\
    \rowcolor{gray! 20} CountGD & $B^{+}_{int}$ & 8.31 & 91.05 & \xmark & \xmark \\
\rowcolor{gray! 20} Ours & $B^{+}_{int}$ & \textbf{6.43} & 32.73 & \xmark & \xmark \\
    C-DETR & $B^{+}_{int}$ & 16.79 & 123.56 & 22.66 & 50.57\\
     PSeCo & $B^{+}_{int}$ & 13.05 & 112.86 & 42.98* & 73.33* \\
  DAVE & $B^{+}_{int}$ & 10.45 & 74.51 & 26.81 & 62.82\\
  \textbf{GeCo} & $B^{+}_{int}$ & 7.91 & 54.28 & \textbf{43.42} & \textbf{75.06}\\
CGD-B & $B^{+}_{int}$ & 10.85 & 99.60 & 34.81 & 69.46\\
Ours & $B^{+}_{int}$ & 8.10 & 35.40 & 38.88 & 73.05 \\
\hline
\multicolumn{6}{c}{Both Text \& Internal (Provided) Exemplars}\\
 \rowcolor{gray!20} CountGD & $t^{+}, B^{+}_{int}$ & \textbf{5.74} & 24.09 & \xmark & \xmark\\
  \rowcolor{gray!20} Ours & $t^{+}, B^{+}_{int}$ & 5.86 & \textbf{18.79} & \xmark & \xmark\\
  CGD-B & $t^{+}, B^{+}_{int}$ & 10.29 & 96.33 & 36.20 & 72.39\\
  \textbf{Ours} & $t^{+}, B^{+}_{int}$ & 7.95	& 29.24 & \textbf{40.29} & \textbf{74.72}\\
\hline
\end{NiceTabular}}
\caption{\label{tab:fscd147_full} Results on \textbf{FSCD-147}~\cite{m_Ranjan-etal-CVPR21, c_detr} for image counting methods in text only, exemplar only, and multi-modal settings. Results for counting methods that do not output boxes are grayed out. Some methods can output the count in multiple ways: by enumerating the output boxes or by applying another approach independent of the output boxes. This is why they appear twice for certain prompt settings. DAVE and DAVE$_{prm}$ output a density map in addition to bounding boxes. The results in gray are for the density-map-based approach, since the objects detected in the density map do not have corresponding boxes. The results in white are from enumerating the output boxes directly. For Ours, the results in gray are from enumerating the bounding boxes and applying test-time normalization, which does not modify the output boxes. The results in white are from just enumerating the bounding boxes. * for PSeCo indicates the result was obtained using the published checkpoints and the same bounding boxes for counting and detection. The symbols for provided prompts are: positive text ($t^{+}$), the 3 manually annotated internal positive visual exemplars from FSC-147 ($B^{+}_{int}$). The abbreviations are: GroundingREC (GrREC~\cite{groundingrec}); CountGD-Box (CGD-B~\cite{countvid}); C-DETR (Counting-DETR~\cite{c_detr}).}
%Results on FSCD-147 Test for image counting methods that output boxes (for Stage~1). `GDINO' is Grounding DINO, `CGD-B' is \countgdbox, and `C-DETR' is Counting-DETR~\cite{10.1007/978-3-031-20044-1_20}. SAM masks are not used in evaluation. `exemp.' means exemplars only, and `both' means both exemplars and text as prompts.}
\end{center}
 
\end{table}

\subsection{Attention Ablation}
In this section, we discuss and ablate our self-attention strategy inside the Feature Enhancer. \methodName\ applies self-attention between positive visual exemplars and text that describe each other and negative visual exemplars and text that describe each other. However, negative prompts describing different classes of objects do not attend to each other, and positive and negative prompts do not attend to each other. We describe different self-attention strategies below.

\begin{figure}[h!]
  \centering
  \includegraphics[width=\linewidth]{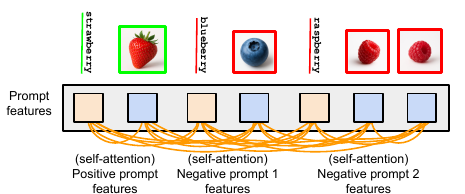}
  \vspace{-3mm}
  \caption{\textbf{[Option A]:} self-attention occurs between all features. In the Feature Enhancer, all the visual exemplar and text prompt features attend to each other regardless of whether they are related.}
  \label{fig:option_a_self_attend}
\end{figure}

\paragraph{[Option A]: All prompts attend to each other.} In this setting, all prompt features attend to each other. This means all positive visual exemplar and text prompts attend to each other, all negative visual exemplar and text prompts attend to each other, and positive prompt features attend to negative ones. In this setting, even if negative concepts are unrelated, dependencies between them are explicitly modeled. An illustration of this strategy is shown in \cref{fig:option_a_self_attend}.

\begin{figure}[h!]
  \centering
  \includegraphics[width=\linewidth]{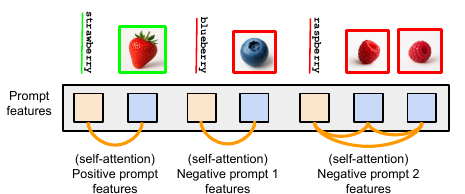}
  \vspace{-3mm}
  \caption{\textbf{[Option B]:} self-attention only occurs between related prompt features. In the Feature Enhancer, corresponding visual exemplar and text features self-attend to each other but not to other visual exemplar and text features. Negative prompts do not attend to each other if they describe different classes.}
  \label{fig:option_b_self_attend}
\end{figure}

\paragraph{[Option B]: Only prompts corresponding to the same concepts attend to each other.}
In this setting, all visual exemplar and text features corresponding to the same class attend to each other. Positive prompt features do not attend to negative prompt features, and negative prompt features that correspond to different classes do not attend to each other. This is the option that we choose as it prevents modeling explicit dependencies between different concepts that may be unrelated. It also does not assume a particular class is positive or negative, allowing this to be chosen after inference has occurred. This means after a single forward pass, all the objects can be counted and subsets of objects can be selected using the model's precomputed output. An illustration of this strategy is shown in \cref{fig:option_b_self_attend}.

In \cref{tab:attention_ablation} we see that Option B, the option that we choose, results in better counting and detection accuracy on the Blood Cell Detection~\cite{blood_cell_detection} dataset. We test on this dataset because it includes both positive and negative visual and textual prompts.

\begin{table}
\begin{center}
{\fontsize{8}{10}\selectfont\begin{NiceTabular}{l|c|c|c|c} 
% \fontsize{9}{11}\selectfont\begin{NiceTabular}{l|l|l|l|l|l|l|c|c|c|c|c|c|c|c} 
\hline

     & \multicolumn{4}{c}{Blood Cell Detection} \\
    
   %\hline
   Attention Strategy & \multicolumn{2}{c}{Counting} & \multicolumn{2}{c}{Detection}\\
   %& Setting & \multicolumn{2}{c}{} & \multicolumn{2}{c}{}\\
   & MAE $\downarrow$ & RMSE $\downarrow$ & AP $\uparrow$ & AP50 $\uparrow$ \\
   \hline
   Option A & 2.03 & 3.13 & 0.40 & 0.66\\
  \textbf{Option B} & \textbf{1.52} & \textbf{2.42} & \textbf{0.54} & \textbf{0.80} \\
   \hline

\end{NiceTabular}}
\vspace{-2mm}
\caption{\label{tab:attention_ablation}Ablation study on the {\bf Blood Cell Detection}~\cite{blood_cell_detection} test set given both positive and negative text and positive and negative external exemplars. In Option A, self-attention is applied between all prompt features in the Feature Enhancer. In Option B, self-attention is only applied between prompt features of the same class.}
\end{center}
 
\end{table}

\subsection{Comparisons to MLLMs}
\begin{figure*}[h!]
  \centering
  \includegraphics[width=\linewidth]{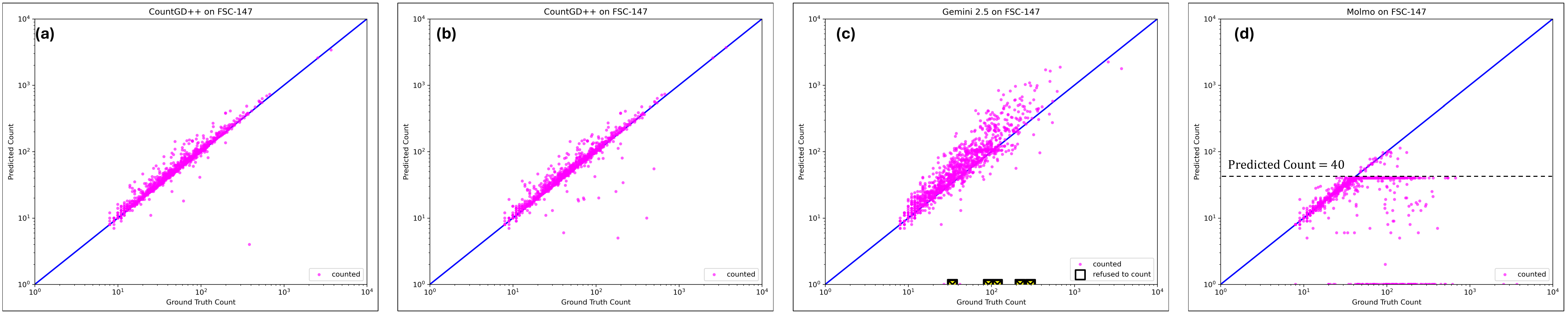}
  \vspace{-3mm}
  \caption{\methodName\ compared to MLLMs on FSC-147 Test. (a) \methodName\ is given both text and the three manually annotated exemplars inside each image; (b) \methodName\ is tested in the case where only text is available while leveraging the pseudo- and synthetic exemplars obtained using only this text; (c) Gemini 2.5~\cite{gemini} is tested on FSC-147 with the prompt ``Count each $class\_name$ and return the total count. Please only provide a single number representing the count. Please be as accurate as possible.'' where $class\_name$ is replaced with the singular form of the class name. Gemini sometimes refuses to count, insisting ``its [my] current capabilities do not allow it [me] to analyze images in that specific way''. These cases are indicated by a box on the $x$-axis; (d) Molmo~\cite{molmo} is tested on FSC-147 with a similar prompt. Molmo counts very poorly after the ground truth count hits 40 as, to avoid memory errors, the model was not trained on data with more than 40 objects to count.
\label{fig:llm_scatter}}
\end{figure*}
In this section, we compare \methodName\ to much larger Multi-modal LLMs (MLLMs) trained on vast quantities of data. We evaluate Gemini-2.5~\cite{gemini} and Molmo~\cite{molmo} on ShanghaiTech Part A~\cite{Zhang_2016_CVPR} and FSC-147~\cite{m_Ranjan-etal-CVPR21} Test and compare their results to ours. For ShanghaiTech, the prompt ``Count the humans in this image. Return only a number.'' is used. For FSC-147, the prompt ``Count each $class\_name$ and return the total count. Please only provide a single number representing the count. Please be as accurate as possible.'' is used. When Gemini~2.5 or Molmo refuses to count, we set the predicted count to 0. Molmo sometimes returns a range instead of a single number, and in these cases we take the midpoint of the range as the estimated count. Despite attempts to improve the prompts for Gemini-2.5 and Molmo, their performance is still significantly worse than \methodName's. These results are presented in \cref{tab:llm_comparison} and \cref{fig:llm_scatter}. This shows specialized counting models still surpass modern MLLMs in counting abilities. We use the Gemini API and model option gemini-2.5-flash-image and the official Molmo GitHub repository and model option Molmo-7B-D to run these experiments. 
\begin{table}
\begin{center}
{\fontsize{8}{10}\selectfont\begin{NiceTabular}{l|c|c|c|c} 
  \hline
   Method & \multicolumn{2}{c}{ShanghaiTech Test Part A} & \multicolumn{2}{c}{FSC-147 Test}\\
   %& Setting & \multicolumn{2}{c}{} & \multicolumn{2}{c}{}\\
   & MAE $\downarrow$ & RMSE $\downarrow$ & MAE $\downarrow$ & RMSE $\downarrow$ \\
   \hline
   Molmo & $7.26 \times 10^{8}$ & $9.47 \times 10^{8}$ & 40.41 & 153.29 \\
  Gemini-2.5 & 517.17 & 1364.40 & 43.18 & 127.02\\
  \textbf{Ours} & \textbf{116.0} & \textbf{234.0} & \textbf{8.39} & \textbf{27.03}\\
\hline
\end{NiceTabular}}
\vspace{-2mm}
\caption{\label{tab:llm_comparison} Comparison with MLLMs. Counting results on Part A of the {\bf ShanghaiTech}~\cite{Zhang_2016_CVPR} crowd counting dataset and the {\bf FSC-147}~\cite{m_Ranjan-etal-CVPR21} dataset given positive text only. Molmo and Gemini-2.5 are compared to \methodName. For ShanghaiTech, \methodName\ uses pseudo exemplars, and for FSC-147, it uses both pseudo- and synthetic exemplars generated from only the text. Molmo performs very poorly on ShanghaiTech compared to both Gemini-2.5 and \methodName, since it has not been trained to count over 40 objects, and ShanghaiTech (A) only has very dense images containing 66+ humans to count. In many cases, Molmo outputs the nonsensical response \emph{1234567890}. Given the maximum number of humans in an image is 2256, such responses increase the average error significantly.}
\end{center}
\end{table}

\begin{figure*}[p]
  \centering
  \includegraphics[width=0.71\linewidth]{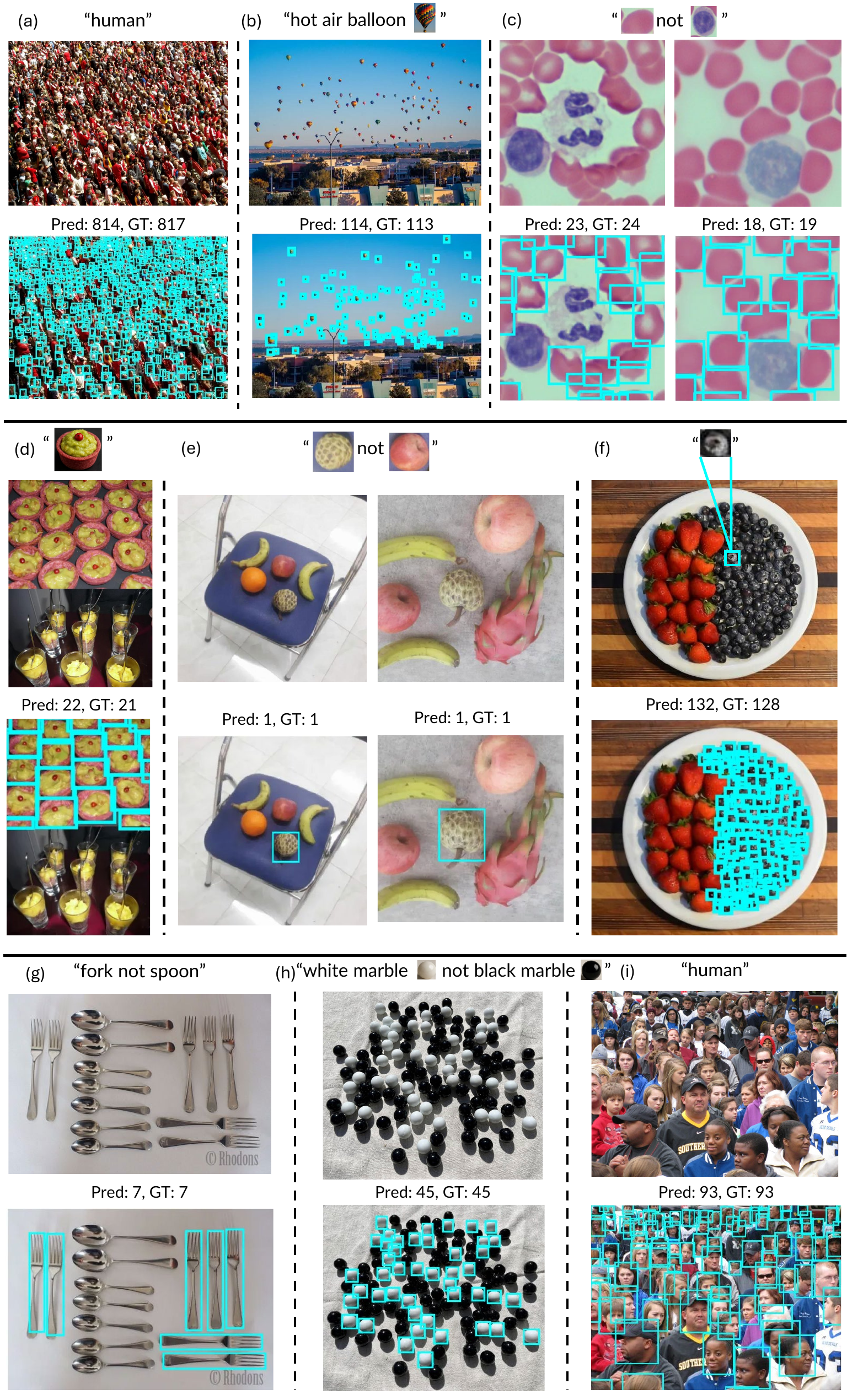}
  \vspace{-3mm}
  \caption{Counting results on images from the test sets and the web. (a), (i) {\bf ShanghaiTech}: positive text and pseudo-exemplars are used to count in dense crowds; (b) {\bf FSCD-147}: positive text, synthetic, and pseudo-exemplars are used. The synthetic exemplar is at the top of the image in the quotes; (c) {\bf Blood Cell Detection}: positive and negative external exemplars from another image are used; (d) {\bf FSCD-147}: two images of different finger foods are mosaicked together, and positive synthetic and pseudo-exemplars are used; (e) {\bf OmniCount}: positive and negative external exemplars are used; (f) {\bf FSCD-147}: one positive manually annotated exemplar is used; (g) {\bf web}: positive and negative text are used; (h) {\bf web}: positive and negative text and positive and negative synthetic exemplars are used.
\label{fig:qual_results_supp}}
\end{figure*}
\begin{figure*}[p]
  \centering
  \includegraphics[width=\linewidth]{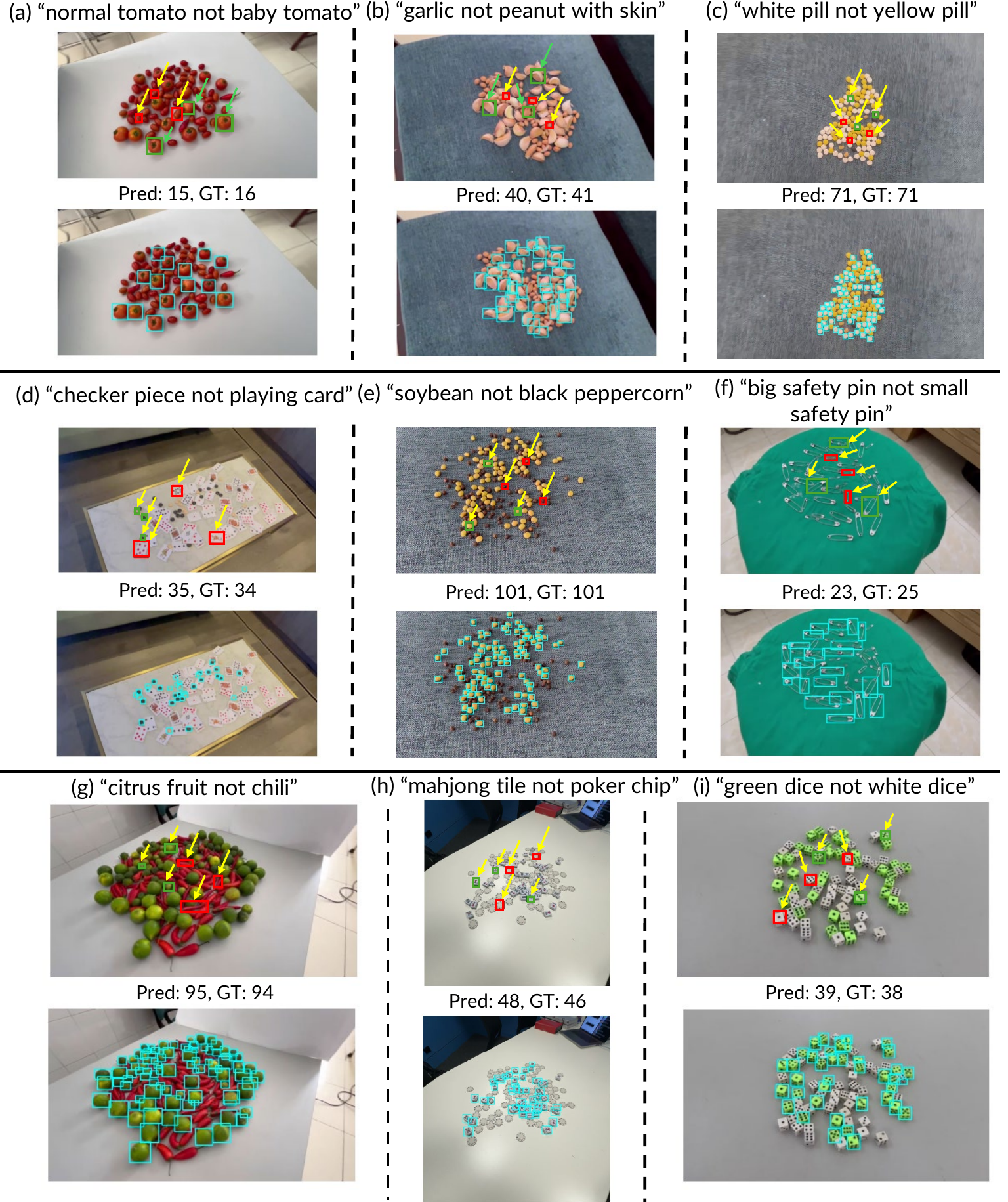}
  \vspace{-3mm}
  \caption{Counting results on images from the \textbf{PairTally} test set given 3 positive and 3 negative exemplars from inside the image and positive and negative text. The positive and negative text are indicated at the top of each image, the positive exemplars are boxed in green and pointed to with the yellow arrows, and the negative exemplars are boxed in red and pointed to with the yellow arrows. \methodName's outputs are shown below the input image and prompts for each example. \methodName\ is able to distinguish between objects with different colors ((c), (e), (g), (i)), the same color but different shapes ((a), (b), (h)), and different sizes ((d), (f)).
\label{fig:qual_results_pairtally_supp}}
\end{figure*}
\section{Further Qualitative Results}
Here we include additional qualitative results from the different aspects of our approach.
\subsection{More \methodName\ Results}
In \cref{fig:qual_results_supp}, we include results from \methodName\ applied to different test images in various prompt settings. \methodName\ is able to count in dense scenes (a, b, i), microscopic out-of-domain images (c), and given only a single synthetic (d) or real (f) exemplar. It can also count given only positive and negative text (g) and differentiate between mixed objects (h).

\subsection{Example Synthetic Exemplar Images}
\begin{figure}[h!]
  \centering
  \includegraphics[width=\linewidth]{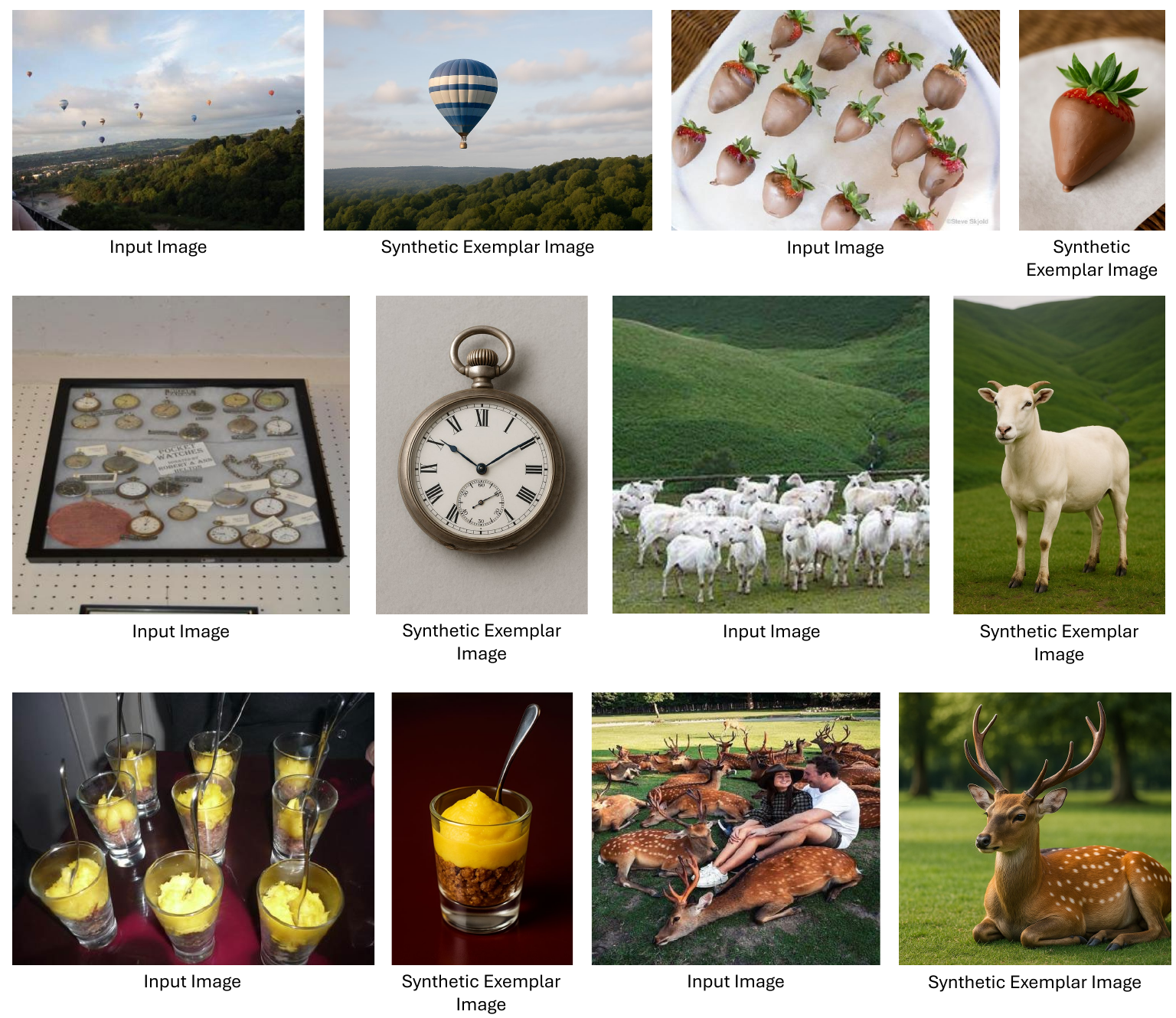}
  \vspace{-3mm}
  \caption{Synthetic exemplar images from our pipeline for generating synthetic exemplars.
\label{fig:synth_exemp_images}}
\end{figure}

In \cref{fig:synth_exemp_images}, we include example synthetic exemplar images generated by GPT-5~\cite{gpt5} in our pipeline for generating synthetic exemplars. Notice how the synthetic exemplar images generally match the style of the input images. This helps \methodName\ match the target object in the input image visually given the synthetic exemplar extracted from the synthetic exemplar image. Also notice that the synthetic images only include one instance. This makes it easier for \methodName\ to crop out the target object to produce the synthetic exemplar.
%\paragraph{HuggingGPT~\cite{hugginggpt} \& \methodName}

\section{Further Implementation Details}
Here we discuss further implementation details about the \methodName\ architecture, training, and inference procedures as well our our synthetic exemplar generation pipeline.
\subsection{Architecture}
\paragraph{Text Encoder. }The text encoder allows for the encoding of multiple object classes in a single forward pass by using a period to distinguish between different concepts. Placing the positive text at the front is the more intuitive choice than placing the positive text at the end. This is because our framework allows for variable numbers of negative concepts, meaning input prompts grow from left to right, as English sentences do. Given the input text prompt containing the positive and negative texts, the text encoder outputs text tokens as 256-dimensional vectors. Notably, the number of vectors representing the texts is determined by the tokenizer, and a single word may correspond to multiple vectors. 

\paragraph{Feature Enhancer. }Before being passed to the Feature Enhancer, the prompt features are first rearranged, so that the positive exemplars follow immediately after the positive text, and the negative exemplars follow immediately after their associated negative text. The ordering for the positives follows directly from~\cite{countgd}, and the ordering for the negatives mirrors this.  

\subsection{Training}
\begin{figure}[h!]
  \centering
  \includegraphics[width=\linewidth]{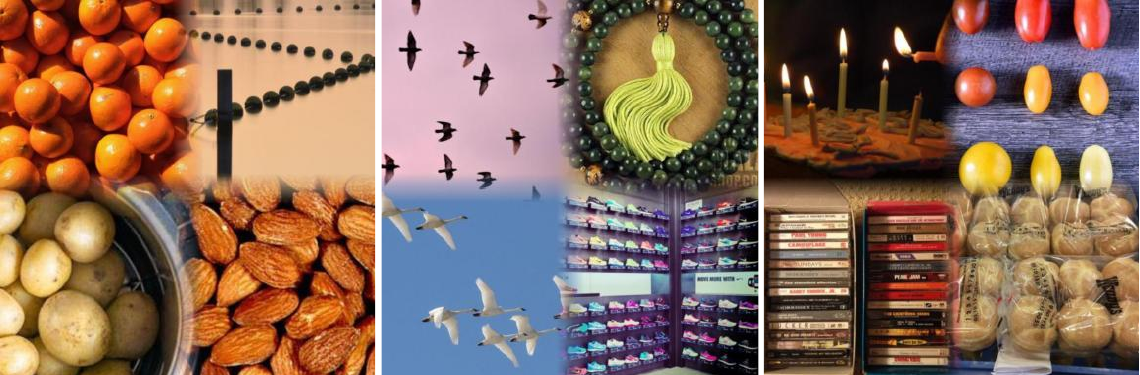}
  \caption{Example synthetic mosaic training images constructed from FSC-147~\cite{m_Ranjan-etal-CVPR21} images. Each image tile of the mosaic provides annotations for a different object category.These mosaic images provide training samples where both positive and negative classes can be specified in the prompt.}
  \label{fig:mosaic}
\end{figure}
To enable specifying negative prompts, we first augment the training set of CountGD-Box~\cite{countvid} with mosaicked images. Examples are shown in \cref{fig:mosaic}. We also modify the input prompts of CountGD-Box~\cite{countvid}. During training, like CountGD-Box, we input a text prompt containing all the training classes, with each class separated by a ``." An example is ``alcohol bottle. baguette roll. ball. banana'' assuming the training set only has these four classes. In practice, the FSC-147~\cite{m_Ranjan-etal-CVPR21} training set we use has 89 categories. For CountGD-Box, only one of the object categories appears in the image at a time, and visual exemplar prompts are only provided for this category. Different from this, we also add visual exemplar prompts for the other categories that appear in the image mosaics, now that they are available. 

The coefficients on the loss terms $\lambda_{loc}$, $\lambda_{GIoU}$, and $\lambda_{cls}$ are set to 5, 2, 2 respectively. These hyperparameters are borrowed directly from CountGD-Box~\cite{countvid} with no further tuning. \methodName\ is trained on FSC-147~\cite{m_Ranjan-etal-CVPR21} with 1000 of our synthetic mosaic images added. The mosaics are constructed by sampling and combining training images uniformly at random using the method in~\cite{Liu22}. The image and text encoders are frozen during training, while the Feature Enhancer and Cross-Modality Interaction are finetuned. To improve the model's ability to count with text only, exemplars only, or both, during training we drop the exemplars 10\% of the time and, if not dropping the exemplars, drop the text 10\% of the time. 

\subsection{Inference}\label{subsec:inference}
At inference, we apply an adaptive cropping procedure to count objects in dense scenes. This is necessary since \methodName\ only has 900 object queries and, thus, can only count up to 900 objects in a single forward pass. To address this, when the count is close to 900, we crop the image into smaller pieces, obtain counts and bounding boxes for each piece individually, and then combine the individual results into a final set of bounding boxes and a final count.

This procedure will now be discussed in detail. When at least 800 objects are counted, the adaptive cropping procedure is activated. Since the model outputs bounding boxes, we can use these to determine the crop height and width to limit the number of objects in each crop. Specifically, we first obtain a set of bounding boxes from the model's output given the whole uncropped image. The minimum object height and width are determined by taking the minimum height and width of the output boxes. The crop width is then set to $25 \times min\_obj\_width$, and the crop height is set to $25 \times min\_obj\_height$. This approximately ensures that at most 625 objects appear in each crop. The factor 25 is chosen as it balances ensuring not too many objects appear in the crop and computational efficiency. A factor too high would risk nearing the model's 900-query limit. A factor too low would require running inference over a high number of crops. The image is cropped into pieces without any overlapping regions. 

\begin{figure}[h!]
  \centering
  \includegraphics[width=\linewidth]{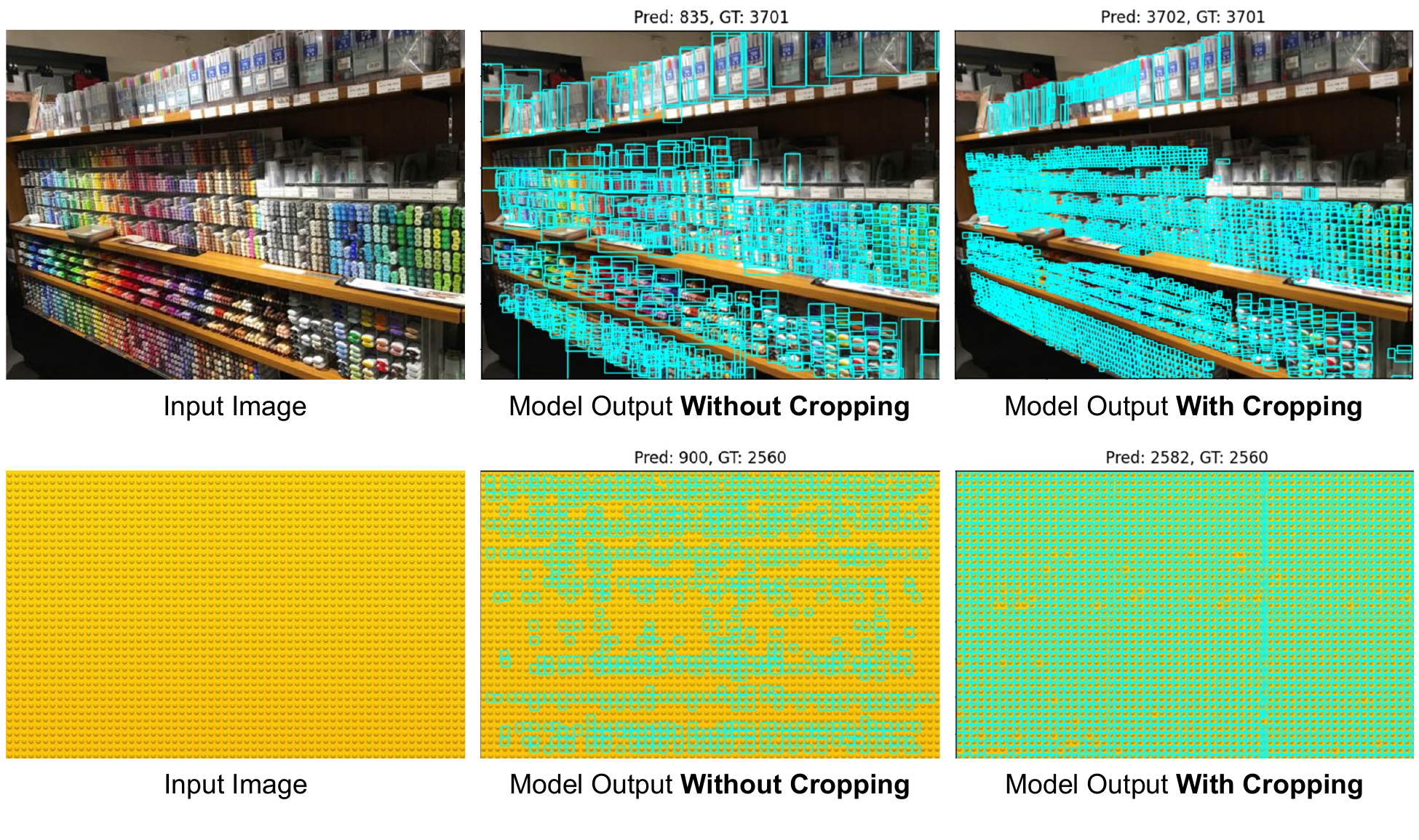}
  \caption{Adaptive cropping improves high-count detection.
\methodName\ can output at most 900 queries per forward pass, so extremely dense scenes cause severe under-counting. Without cropping (middle column), the model either merges many nearby objects (markers in top row) or misses large numbers of instances (yellow lego studs in bottom row). With adaptive cropping (right column), each crop remains within the model’s capacity, allowing it to detect far more objects. As a result, more instances are picked up and bounding boxes less frequently merge multiple objects.}
  \label{fig:adaptive_cropping_results}
\end{figure}

\begin{figure}[h!]
  \centering
  \includegraphics[width=\linewidth]{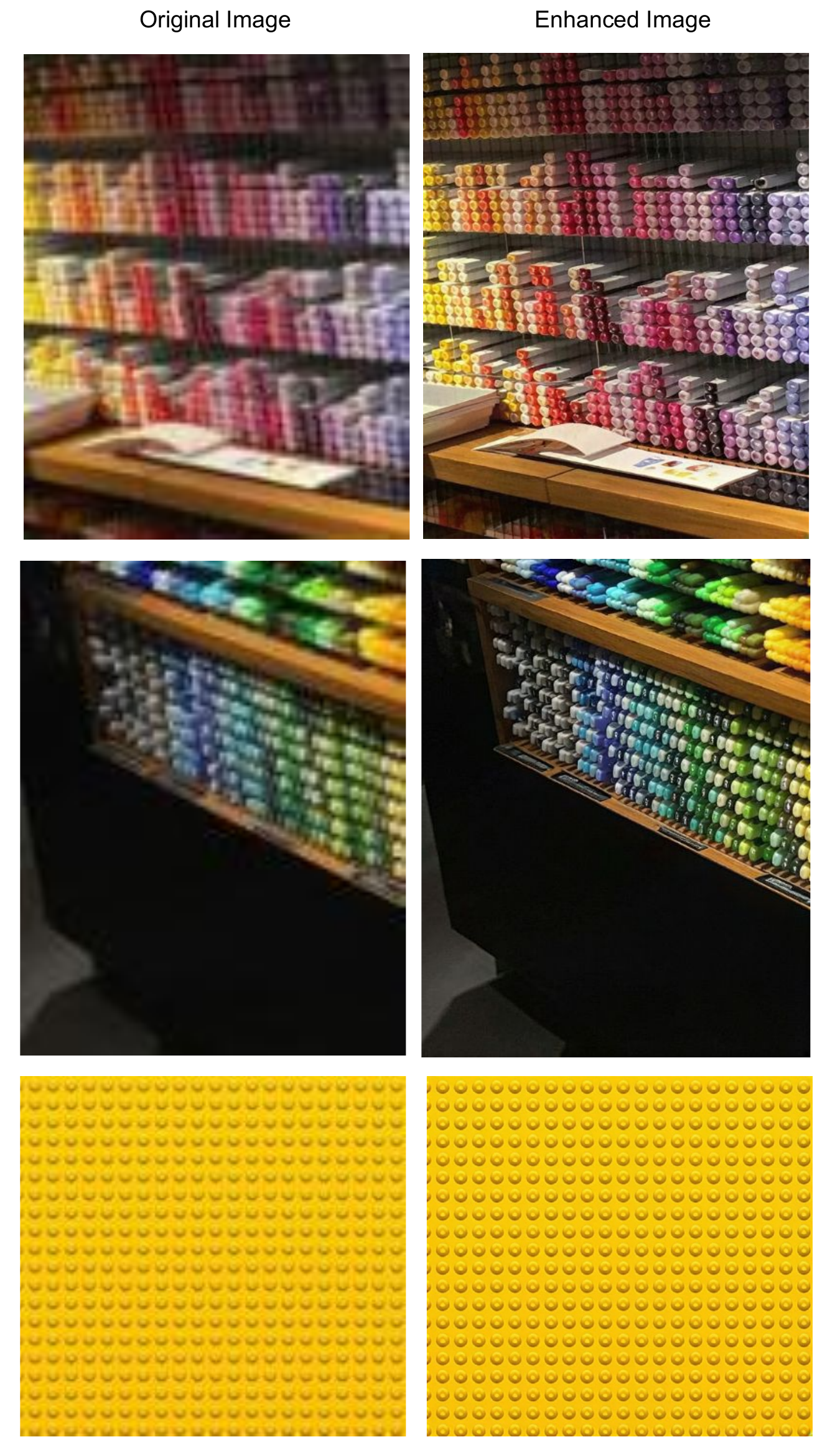}
  \caption{Super-resolution enhances crops for adaptive cropping. In our adaptive cropping procedure, we apply the super-resolution method in \cite{bubbi_app} to the cropped images to help \methodName\ pick out the objects. Note how the super-resolution does not change the count or locations of the objects.}
  \label{fig:adaptive_cropping_super}
\end{figure}

Because in dense scenes the objects appear small, they can be difficult and too blurry for the model to pick out in the crops. To address this, we apply super-resolution to upscale the image crops by a factor of 4 with the Standard AI Image Upscaler from \cite{bubbi_app}. This method is chosen because it preserves the count and locations of the objects in the crops, so bounding boxes and counts can be obtained. The final set of bounding boxes is the union of the bounding boxes from the crops and the final count is the sum of the counts for the crops. 
\begin{comment}
\az{I would remove from here on -- it is too detailed, and probably will change/improve ..} For the cropped images, only text is used as we find that using real or synthetic exemplars sometimes induces false positives as the high-resolution images may look different from the original low resolution ones. If text is not available, the exemplars are used anyway. Note, there is no special treatment of the boundaries, so false positives may occur from the same object being detected along boundaries that meet from different crops. However, because adaptive cropping only occurs in dense scenes, the errors from these false positives are negligible relative to the number of instances.

\az{I would omit the following paragraph.}
We did not perform extensive hyperparameter tuning for the adaptive cropping procedure. This is primarily because the super-resolution step relies on a commercial image enhancement tool (the Bubbi App~\cite{bubbi_app}), and each enhancement incurs a monetary cost. Exhaustively exploring crop parameters would therefore be expensive. Despite this, we found that a single heuristic setting (crop dimension factor = 25, upscale factor = 4) was sufficient and robust across all cases where cropping was actually triggered. The enhanced image crops will be publicly released.

\end{comment}
In \cref{fig:adaptive_cropping_results} we show examples of the original model predictions, and the predictions after applying our adaptive cropping procedure. In \cref{fig:adaptive_cropping_super} we show examples of the original crops and the same crops enhanced with super-resolution.


\subsection{Synthetic Exemplar Generation}
Here we include the prompt template provided to GPT-5~\cite{gpt5} for generating the synthetic exemplar images. For almost all the classes in FSC-147~\cite{m_Ranjan-etal-CVPR21} Test, we use the template ``generate an image of a single $class\_name$ in the reference image. Please make the instance of the $class\_name$ match the $class\_name$s in the reference image as closely as possible." where $class\_name$ is replaced with the class name. For the ``stamp" and ``comic book" classes, we use a modified version of this prompt to avoid copyright issues that trigger GPT-5's safety guardrails. The modified prompt template is ``generate an image of a $class\_name$ in the same style as the ones in the reference image.'' Example synthetic exemplar images are included in \cref{fig:synth_exemp_images}. The synthetic exemplar images generated for FSC-147 Test will be publicly released.

\section{Further Dataset Details}

Here we include detailed information about our different datasets. We also discuss why certain datasets were omitted.

\paragraph{FSCD-147~\cite{m_Ranjan-etal-CVPR21, c_detr}. } FSCD-147, adds bounding boxes to the validation and test sets of FSC-147, the standard dataset for open-world counting, containing 6135 images with 89 classes in the training set, 29 classes in the validation set, and 29 classes in the test set. The classes in the different sets are disjoint. Each image is annotated with 3 visual exemplars. Instances of only a single object class are labeled per image, and each image has 7-3731 objects. Bounding boxes are provided for the validation and test sets.

\paragraph{PrACo~\cite{mind_the_prompt}. } PrACo is a counting benchmark constructed from images in FSCD-147. It introduces the Negative Label Test to evaluate counting models when the target object is not in the image and the Mosaic Test to evaluate counting models in the multi-class setting.

\paragraph{ShanghaiTech~\cite{Zhang_2016_CVPR}. } ShanghaiTech is a crowd counting dataset composed of Part A, with 182 test images containing 66-2256 humans per image, and Part B, with 316 images containing 9-539 humans per image. Each human is annotated with a dot.

\paragraph{Blood Cell Detection~\cite{blood_cell_detection}. } This dataset contains exhaustive bounding box annotations for red and white blood cells in 100 images from a peripheral blood smear taken from a light microscope. A peripheral blood smear is a technique for microscopic blood cell examination and can aid in medical diagnosis. The images contain 11-33 cells each.

\paragraph{OmniCount (Fruits)~\cite{omnicount}. } This is the Fruits test set of the OmniCount-191 benchmark. It contains 303 images with 8 different fruits annotated with bounding boxes. Each image contains 3-6 fruits. The other test sets of OmniCount-191 were omitted for reasons detailed in \cref{sec:omni_omit}.

\paragraph{VideoCount (Crystals)~\cite{countvid}. } This is the Science-Count (Crystals) test set of the VideoCount benchmark. It contains 10 videos of 10-154 crystals rapidly forming from liquid metal alloys in x-ray videos. The number of unique crystals in each video is annotated. We choose this test set of VideoCount, since the crystals significantly change size and structure over time, demonstrating the benefit of dynamic pseudo-exemplars.

\paragraph{PairTally~\cite{nguyen2025pairtally}. }The PairTally dataset contains 681 images that test a counting model's ability to distinguish between different objects within an image. The dataset includes cases where there are only subtle differences in the the shape, color, texture, and size of different objects. There are usually many instances of each object type mixed together and placed on a surface. The camera angle varies and sometimes introduces challenging perspective effects. Most images contain many instances, with over 150 images with 200+ total instances. PairTally has been shown to be a very challenging dataset for counting models. The metadata and ground truth annotations include text prompts, 3 exemplars, and center points for all the objects to be counted.

\paragraph{CARPK~\cite{Hsieh2017DroneBasedOC}. }

The Car Parking Lot Dataset (CARPK) contains nearly 90,000 cars from 4 different parking lots captured via a drone from about 40 meters away. Each car is annotated with a bounding bounding box. There are 989 training images with 1-87 cars per image and 459 test images with 2-188 cars per image. 

\paragraph{Omission of the Rest of OmniCount-191~\cite{omnicount} .}\label{sec:omni_omit}
At the time of our experiments, the publicly released version of OmniCount-191 exhibited several issues that prevented reliable evaluation on most of its subsets. We observed that ground-truth boxes for some classes (e.g., birds) were missing, aspect ratios for other classes (e.g., pets) were distorted, and in some cases objects that should be counted as distinct units were grouped into a single box (e.g., multiple houses annotated as one instance). In contrast, the Fruits subset contained accurate annotations and included uncommon fruit categories (e.g., sugar apple) that provide valuable test cases for open-world counting. For these reasons, we restricted our evaluation to the Fruits subset.

\begin{comment}
\az{I would omit the following paragraph -- we can use this argument in the rebuttal if a reviewer asks about REC-8k.}
\paragraph{Omission of REC-8k~\cite{groundingrec} .}
REC-8k~\cite{groundingrec} is a dataset for Referring Expression Counting, where the goal is to count objects specified by a natural-language referring expression (e.g., “the books on the top shelf,” “the people sitting down”). REC-8k does not provide visual exemplars or bounding box annotations. Furthermore, unlike our setting, the text in Referring Expression Counting does not represent a class name but a description of an object, its attributes, and its context (such as location). Because our method relies on class-level text prompts and benefits from internal or external visual exemplars, REC-8k is not directly compatible with our problem formulation. Therefore, we do not train or evaluate on this dataset.
\end{comment}
\section{Further Clarifications}
\subsection{PSeCo~\cite{zhizhong2024point} Evaluation}
The evaluation protocol used in PSeCo differs from the standard one in the counting literature~\cite{Pelhan_2024_NeurIPS, dave, countvid, c_detr}. In PSeCo, the bounding boxes used for counting are not the same as the ones used for detection, whereas in the standard protocol a single set of boxes is used for both tasks. This discrepancy has also been noted by prior work~\cite{Pelhan_2024_NeurIPS}. To ensure consistent comparison across methods, in Tab. 1 of the main paper and \cref{tab:fscd147_full} of the supplementary we report PSeCo’s results recomputed using the standard evaluation protocol, applying the same set of boxes for both counting and detection for all methods.

\subsection{Prompting With Negatives}
Beyond the counting literature, the use of negatives in related fields is more prominent. For detecting in out-of-distribution images, NegPrompt~\cite{neg_prompt} learns negative prompts using positive class labels. Unlike our approach, the negative prompts cannot be provided explicitly at inference and are instead learned implicitly from the positive prompts. 
In segmentation and tracking, SAM~2~\cite{ravi2024sam2} allows users to specify \emph{negative clicks} identifying regions that should not be segmented at inference. Similarly, in retrieval, users can provide negative prompts as feedback to refine model predictions~\cite{user_feedback_img_retr, neg_prompt_retr}. Negative prompting has also been widely explored in text-to-image generation methods such as Stable Diffusion~\cite{ban2024understanding}; however, despite its success there, understanding explicit negation remains challenging for VLMs like CLIP~\cite{neg_bench, Radford2021LearningTV}.

\subsection{Open-World vs. Open-Vocabulary}
Here we make an important note about terminology. We notice that in the object counting literature~\cite{AminiNaieni23, countgd, Liu22}, {\em open-world counting} refers to the task of counting instances of an object class specified at test time via textual or visual prompts. These counting models are {\em open-world} because they generalize beyond a fixed vocabulary, as the object of interest may belong to a class not seen during training. However, crucially, the category is still explicitly provided as input at inference time, either as text or exemplar. We adopt this definition of {\em open-world} in our work, since \methodName\ builds on counting literature.

However, this usage of {\em open-world} differs from that in some earlier literature. The origins of the formal definition of `open world' are from \cite{owr}. This paper defines an open-world recognition system as one that recognizes instances of both \emph{known} and \emph{unknown} classes, marks the unknown objects as `unknown,' obtains class labels for these unknown objects, and incrementally learns from these instances such that they become `known.' Similarly, early works in open-world object detection \cite{joseph2021open} aim to discover and learn novel object categories without labels at first, marking them as `unknown,' and incrementally learning them later. However, more recent works in detection~\cite{owlv2, owl} use the term `open-world' to describe a detection model that can detect objects unseen during training by accepting textual prompts (i.e., labels) describing the object. Importantly, this means that the labels of the unknown objects must be provided for the model to detect them, which differs from the original formal definition of `open world' in \cite{owr}. The OWL paper~\cite{owl} uses the terms `open-world' and `open-vocabulary interchangeably to describe this prompt-based setting as we do. In fact `OWL' stands for `Open-World Localization.' Like OWL, in our case, the category is always specified by the user, and the challenge lies in handling domain shifts, visual diversity, and lack of training-time exposure to the category.

% WARNING: do not forget to delete the supplementary pages from your submission 
%\input{sec/X_supp}

\end{document}